\documentclass[sigconf]{acmart}
\usepackage{algorithmic}
\usepackage{multirow}
\usepackage{subcaption}
\usepackage{enumitem}
\usepackage{graphicx}
\usepackage[ruled]{algorithm2e}

\newlength\savedwidth

\newlength\savewidth
\newcommand\shline{\noalign{\global\savewidth\arrayrulewidth
                            \global\arrayrulewidth 1.5pt}%
                   \hline
                   \noalign{\global\arrayrulewidth\savewidth}
                   }

\def\BibTeX{{\rm B\kern-.05em{\sc i\kern-.025em b}\kern-.08em
    T\kern-.1667em\lower.7ex\hbox{E}\kern-.125emX}}

\makeatletter
\newcommand{\thickhline}{%
    \noalign {\ifnum 0=`}\fi \hrule height 1pt
    \futurelet \reserved@a \@xhline
}

\newcommand\blfootnote[1]{%
  \begingroup
  \renewcommand\thefootnote{}\footnote{#1}%
  \addtocounter{footnote}{-1}%
  \endgroup
}

\AtBeginDocument{%
  \providecommand\BibTeX{{%
    \normalfont B\kern-0.5em{\scshape i\kern-0.25em b}\kern-0.8em\TeX}}}

\copyrightyear{2022} 
\acmYear{2022}
\setcopyright{rightsretained} 
\acmConference[SIGSPATIAL '22]{The 30th International Conference on Advances in Geographic Information Systems}{November 1--4, 2022}{Seattle, WA, USA}
\acmBooktitle{The 30th International Conference on Advances in Geographic Information Systems (SIGSPATIAL '22), November 1--4, 2022, Seattle, WA, USA}
\acmDOI{10.1145/3557915.3560939}
\acmISBN{978-1-4503-9529-8/22/11}

\begin{document}

\title{When Do Contrastive Learning Signals Help Spatio-Temporal Graph Forecasting?}


\author{Xu Liu$^{1*}$, Yuxuan Liang$^{1*}$, Chao Huang$^2$, Yu Zheng$^{3,4}$, Bryan Hooi$^1$, Roger Zimmermann$^1$
}

\affiliation{$^1$School of Computing, National University of Singapore \country{Singapore} \\ 
    $^2$Department of Computer Science, University of Hong Kong \country{China} \\ 
    $^3$JD Intelligent Cities Research \country{China} \quad $^4$JD iCity, JD Technology \country{China} \\
    \{liuxu, yuxliang, bhooi, rogerz\}@comp.nus.edu.sg; chaohuang75@gmail.com; msyuzheng@outlook.com
}

\renewcommand{\shortauthors}{Liu et al.}

%%
%% The abstract is a short summary of the work to be presented in the article.
\begin{abstract}
    Deep learning models are modern tools for spatio-temporal graph (STG) forecasting. Though successful, we argue that data scarcity is a key factor limiting their recent improvements. Meanwhile, contrastive learning has been an effective method for providing self-supervision signals and addressing data scarcity in various domains. In view of this, one may ask: can we leverage the additional signals from contrastive learning to alleviate data scarcity, so as to benefit STG forecasting? To answer this question, we present the first systematic exploration on incorporating contrastive learning into STG forecasting. Specifically, we first elaborate two potential schemes for integrating contrastive learning. We then propose two feasible and efficient designs of contrastive tasks that are performed on the node or graph level. The empirical study on STG benchmarks demonstrates that integrating graph-level contrast with the joint learning scheme achieves the best performance. In addition, we introduce four augmentations for STG data, which perturb the data in terms of graph structure, time domain, and frequency domain. Experimental results reveal that the model is not sensitive to the proposed augmentations' semantics. Lastly, we extend the classic contrastive loss via a rule-based strategy that filters out the most semantically similar negatives, yielding performance gains. We also provide explanations and insights based on the above experimental findings. Code is available at https://github.com/liuxu77/STGCL. \blfootnote{*The first two authors contributed equally to this work.}
\end{abstract}
% Meanwhile, contrastive learning as a surging self-supervised learning techniques have been applied in a wide range of domains, e.g., the graph domain, for learning more transferrable, generalizable, and robust representations.
% The consistent improvements demonstrate the effectiveness of STGCL and its generalization ability to existing spatio-temporal neural networks.
% they require large-scale datasets to achieve better performance and are vulnerable to noise perturbations. 
% The consistent improvements demonstrate that STGCL can be used as an off-the-shelf plug-in for existing spatio-temporal neural networks.
% Spatio-temporal graph forecasting is of great importance to a wide range of real-world applications, such as traffic prediction. Despite the success of designing models to capture complex spatio-temporal dependencies, we argue that this task has two inherent limitations found in common real-world datasets: 1) limited training samples can be constructed; and 2) the data collected from sensors are not perfectly accurate and usually contain noise.

\begin{CCSXML}
<ccs2012>
<concept>
<concept_id>10002951.10003227.10003236</concept_id>
<concept_desc>Information systems~Spatial-temporal systems</concept_desc>
<concept_significance>500</concept_significance>
</concept>
</ccs2012>
\end{CCSXML}
\ccsdesc[500]{Information systems~Spatial-temporal systems}

%%
%% Keywords. The author(s) should pick words that accurately describe the work being presented. Separate the keywords with commas.
\keywords{Spatio-temporal graph, contrastive learning, deep neural network}

\maketitle

\begin{figure}[!h]
  \centering
  \includegraphics[width=\linewidth]{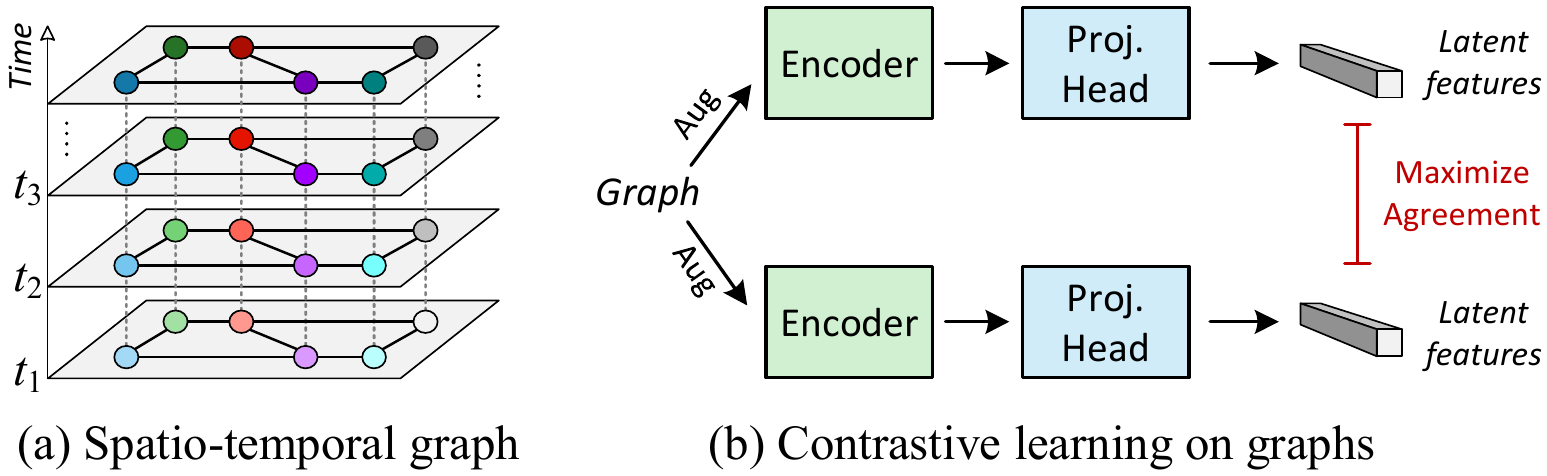}
  \vspace{-1.5em}
  \caption{Illustration of STG and contrastive learning. The latent features can be either node \cite{zhu2021graph} or graph \cite{you2020graph} features. Aug: augmentation. Proj: projection.}
  \label{fig:intro}
  \vspace{-1em}
\end{figure}

\section{Introduction}
\label{sec:1}
Deploying a large number of sensors to perceive an urban environment is the basis for building a smart city. The time-varying data that are produced from the distributed sensors can usually be represented as a spatio-temporal graph (STG), as shown in Figure \ref{fig:intro}(a). Leveraging the collected data, one important application is to forecast future trends based on historical observations, such as traffic forecasting. State-of-the-art approaches for this problem typically use convolutional neural networks (CNN) \cite{wu2019graph, guo2019attention, song2020spatial, li2021spatial, han2021dynamic} or recurrent neural networks (RNN) \cite{li2018diffusion, pan2019urban, bai2020adaptive, cao2020spectral} to model temporal dependencies. For capturing spatial correlations, these methods mostly utilize the popular Graph Neural Networks (GNN) \cite{kipf2017semi, velivckovic2018graph}.

While tremendous efforts have been made to design sophisticated architectures to capture complex spatio-temporal correlations, we argue that \emph{data scarcity} is an essential issue that hinders the recent improvements on STG forecasting. Specifically, public datasets in this area usually possess only a few months of data, restricting the number of training instances that can be constructed. For example, the frequently used benchmarks PEMS-04 and PEMS-08 \cite{guo2019attention} only have around 17,000 instances in total, which is relatively limited compared to the datasets in other domains, such as images and text. Consequently, the learned models may overfit the training data, leading to inferior generalization performance.
% Second, the sensor readings are never perfectly accurate or sometimes missing due to unexpected factors, such as signal interruptions \cite{yi2016st}. Unfortunately, most existing models tend to be vulnerable to these factors and their robustness need to be improved.

Meanwhile, self-supervised learning has demonstrated great promise in a series of tasks on graphs. It derives supervisory signals from the data itself, usually exploiting the underlying structure of the data. Most of the best performing self-supervised methods are based on \emph{contrastive learning} \cite{velickovic2019deep, qiu2020gcc, you2020graph, hassani2020contrastive}. As depicted in Figure \ref{fig:intro}(b), their basic idea is to maximize the agreement between representations of nodes or graphs with similar semantics (i.e., positive pairs), while minimizing those with unrelated semantic information (i.e., negative pairs). Generally, positive pairs are established by applying data augmentations to generate two views of the same input (termed \textit{anchor}) \cite{you2020graph, zhu2021graph}, and the negative pairs are formed between the anchor and \emph{all} other inputs' views within a batch. Moreover, there are two schemes to utilize the contrastive learning signals, i.e., pretraining \& fine-tuning \cite{qiu2020gcc, hu2020gpt} or joint learning \cite{you2020does, zeng2021contrastive}.

% Generalizable and robust representations are obtained from unlabeled data

% Witnessing recent advances in self-supervised representation learning on graphs \cite{you2020does, velickovic2019deep, qiu2020gcc, you2020graph}, we argue that contrastive learning is a natural fit to address these two issues. First, data augmentation, which is a crucial component in contrastive learning technique, can not only enlarge the limited datasets but also considerably reduce the impact of noise. Second, a contrastive loss is proposed to regularize the derived representations of nodes/graphs. It maximizes the agreement between representations of nodes or graphs generated by augmentation, while minimizing those with unrelated semantic information. In this way, models are trained to be aware and tailored the noise.

\textbf{Contribution.}
In light of the success of contrastive learning, we present the first systematic study to answer a key question: \emph{can we leverage the additional self-supervision signals derived from contrastive learning techniques to alleviate data scarcity, so as to benefit STG forecasting?} If so, how should we integrate contrastive learning into spatio-temporal neural networks? Concretely, we identify the following questions to address.
\begin{enumerate}[leftmargin=*,label={\textbf{Q\arabic*}:}]
    \item What is the appropriate training scheme when integrating contrastive learning with STG forecasting?
    \item For STG, there are two possible objects (node or graph) to conduct contrastive learning. Which one should we select?
    \item How should we perform data augmentation to generate a positive pair? Does the choice of augmentation methods matter?
    \item Given an anchor, should all other objects be considered as negatives? If not, how should we filter out unsuitable negatives?
\end{enumerate}

In this paper, we give an affirmative answer by presenting a novel framework, entitled \emph{STGCL}, which incorporates the merits of contrastive learning into spatio-temporal neural networks. The framework can be regarded as a handy training strategy to improve performance without extra computational cost during inference. We also provide explanations and insights that derive from the empirical findings. The above questions are answered as follows.
% We summarize our contributions by addressing the above questions.
\begin{enumerate}[leftmargin=*,label={\textbf{A\arabic*}:}]
    \item We demonstrate the effectiveness of jointly conducting the forecasting and contrastive tasks over the pretraining scheme through evaluations on popular STG benchmarks.
    
    \item We explore feasible and efficient designs of contrastive learning on two separate objects -- node and graph level. The empirical study shows that integrating graph-level contrast with the joint learning scheme achieves the best performance.
    
    \item We introduce four types of data augmentation methods for STG data which perturb the inputs in three aspects -- graph structure, time domain, and frequency domain. Our experiments reveal that the model is not sensitive to the semantics of the proposed augmentations.
    
    \item We propose to filter out the hardest negatives per anchor based on the unique characteristics of STG, e.g., temporal closeness\footnote{Temporal closeness means time intervals in the recent time \protect \cite{zhang2017deep,guo2019attention,liang2021fine}} and periodicity, yielding improvements on the prediction accuracy over real-world benchmarks.
\end{enumerate}

\section{Preliminaries}
\subsection{Spatio-Temporal Graph Forecasting}
We define a sensor network as a graph ${G} = (\mathcal{V}, \mathcal{E}, \mathbf{A})$, where $\mathcal{V}$ is a set of nodes and $|\mathcal{V}| = N$, $\mathcal{E}$ is a set of edges, and $\mathbf{A} \in \mathbb{R}^{N \times N}$ is a weighted adjacency matrix where the edge weights represent the proximity between nodes. The graph $G$ has a unique feature matrix $\mathbf{X}^t \in \mathbb{R}^{N \times F}$ at each time step $t$, where $F$ is the feature dimension. The features consist of a target attribute (e.g., traffic speed or flow) and other auxiliary information, such as time of day \cite{wu2019graph}. In STG forecasting, we aim to learn a function $f$ to predict the target attribute of the next $T$ steps based on $S$ historical frames:
\begin{equation}
    \mathcal{G}: [\mathbf{X}^{(t-S):t}; G] \xrightarrow{f} \mathbf{Y}^{t:(t+T)}
\end{equation}
where $\mathbf{X}^{(t-S):t} \in \mathbb{R}^{S \times N \times F}$ indicates the observations from the time step $t-S$ to $t$ (excluding the right endpoint), and $\mathbf{Y}^{t:(t+T)} \in \mathbb{R}^{T \times N \times 1}$ denotes the $T$-step ahead predictions.

\subsection{Contrastive Learning on General Graphs}
Recent studies have verified the power of contrastive learning in learning unsupervised representations of graph data \cite{you2020graph,zhu2021graph}. Generally, the contrastive object varies with the downstream tasks, e.g., node-level contrastive learning is applied to learn useful representations before performing node classification tasks \cite{zhu2021graph}.

The pipeline of node- and graph-level contrastive learning can be summarized into a unified form (as illustrated in Figure \ref{fig:intro}(b)). Firstly, given an input graph, we utilize data augmentation methods to generate two correlated views. These views then pass through a GNN encoder to obtain the corresponding node or graph (derived from a readout function) representations. A projection head, i.e., non-linear transformations, further maps the representations to another latent space, where the contrastive loss is computed.

% Let $\mathbf{z}_{i}^{\prime}$ and $\mathbf{z}_{i}^{\prime\prime}$ denote the two correlated views' projection head outputs from the $i$th node/graph in a batch
During training, a batch of $M$ nodes/graphs are sampled and processed through the above procedure, resulting in $2M$ representations in total. Let $\mathbf{z}_{i}^{\prime}$ and $\mathbf{z}_{i}^{\prime\prime}$ denote the latent features (i.e., the outputs of the projection head) of the first and second views of the $i$th node/graph, and ${\rm sim}(\mathbf{z}_{i}^{\prime}, \mathbf{z}_{i}^{\prime\prime})$ denote the cosine similarity between them. Below shows an example of the contrastive loss applied in GraphCL \cite{you2020graph}, which is a variant of the InfoNCE loss \cite{oord2018representation, chen2020simple} and computed across the nodes/graphs in a batch:
\begin{equation}
    \mathcal{L}_{cl} = \frac{1}{M} \sum_{i=1}^{M} - \log \frac{\exp({\rm sim}(\mathbf{z}_{i}^{\prime}, \mathbf{z}_{i}^{\prime\prime}) / \tau)}{\sum_{j=1, j\neq i}^{M} \exp({\rm sim}(\mathbf{z}_{i}^{\prime}, \mathbf{z}_{j}^{\prime\prime}) / \tau)}
    \label{eq1}
\end{equation}
where $\tau$ denotes the temperature parameter and a total of $M - 1$ negatives are incorporated for the $i$th node/graph. After pretraining, we leverage the learned representations to perform downstream tasks, e.g, node or graph classification.
% To use the pretrained model for downstream tasks, such as node/graph classifications, a linear classifier is trained using cross-entropy loss on top of the GNN encoder, which could be frozen (linear evaluation) or unfrozen (fine-tuning). The projection head is discarded.

\subsection{STG Encoder \& Decoder}
\label{sec:2.3}
Based on the techniques exploited in the temporal dimension, existing STG forecasting models mainly fall into two classes: CNN-based and RNN-based methods. Here, we briefly introduce their architectures, which are generally in a encoder-decoder fashion.

For the form of the STG encoder $f_{\theta}(\cdot)$, CNN-based methods usually apply temporal convolutions and graph convolutions sequentially \cite{wu2019graph} or synchronously \cite{song2020spatial} to capture spatio-temporal dependencies. The widely applied form of temporal convolution is the dilated causal convolution \cite{yu2016multi}, which enjoys an exponential growth of the receptive field by increasing the layer depth. After several layers of convolution, the temporal dimension is eliminated because the knowledge has been encoded into the representations. While in RNN-based methods, graph convolution is integrated with recurrent neural networks \cite{li2018diffusion}. The hidden representation from the last step is usually seen as a summary of the encoder. In short, both CNN- and RNN-based models' encoder output
\begin{equation}
    \mathbf{H} = f_{\theta}([\mathbf{X}^{(t-S):t}; G])
\end{equation}
where $\mathbf{H} \in \mathbb{R}^{N \times D}$, and $D$ is the size of the hidden dimension.

For the form of the STG decoder $g_{\phi}(\cdot)$, CNN-based models often apply several linear layers to map high dimensional representations to low dimensional outputs \cite{wu2019graph}. In RNN-based models, they either employ a feed-forward network \cite{bai2020adaptive} or a recurrent neural network \cite{li2018diffusion} for generating the forecasting results. Lastly, a prediction loss $\mathcal{L}_{pred}$, e.g., mean absolute error (MAE), is utilized to train the neural networks. Formally, we have
\begin{equation}
    \mathcal{L}_{pred} =  |\hat{\mathbf{Y}}^{t:(t+T)} - \mathbf{Y}^{t:(t+T)}|, \text{ where } \hat{\mathbf{Y}}^{t:(t+T)} = g_{\phi}(\mathbf{H}).
\end{equation}

% For brevity, some frequently used notations are listed in Table \ref{tab:notations}.
% \begin{table}[!h]
%   \centering
%   \small
%   \caption{Frequently used notations.}
%   \tabcolsep=3mm
%   \vspace{-0.5em}
%     \begin{tabular}{c|c}
%     \hline
%     \textbf{Notations} & \textbf{Description} \\
%     \hline
%       $\mathbf{X}^t$   & The feature matrix at time $t$ \\
%       $\mathbf{Y}^t$   & The target for prediction at time $t$ \\
%       $f_\theta,g_\phi$ & The function of STG encoder/decoder \\
%       $\mathbf{z}_{i}^{\prime},\mathbf{z}_{i}^{\prime\prime}$ & First, second view's features of the $i$th node/graph \\
%       $\mathcal{L}_{pred},\mathcal{L}_{cl}$   & Loss function of prediction/contrastive learning \\
%      \hline
%     \end{tabular}%
%   \label{tab:notations}%
% \end{table}%

\section{Methodology}
In this section, we will introduce the proposed STGCL framework by addressing the four essential questions. We start by elaborating two possible schemes to incorporate contrastive learning with STG forecasting in Section \ref{sec:3.1}. We then describe two designs of contrastive tasks on different levels (node and graph level) in Section \ref{sec:3.2}. Afterwards, we introduce four kinds of data augmentations for STG data in Section \ref{sec:3.3} and illustrate a rule-based strategy to exclude undesirable negatives when calculating contrastive loss in Section \ref{sec:3.4}. Finally, we link all the proposed techniques and provide a sample implementation in Section \ref{sec:3.5}.

\subsection{Training Schemes (Q1)}
\label{sec:3.1}
In this part, we discuss two candidate schemes to incorporate contrastive learning with STG forecasting.

\subsubsection{Pretraining \& fine-tuning.}
Following previous works \cite{you2020graph, zhu2021graph}, we first generate the first view $\mathcal{G}^{\prime}$ and the second view $\mathcal{G}^{\prime\prime}$ of the same input $\mathcal{G}$ by augmentation (see details in Section \ref{sec:3.3}). Then, we feed the views through an STG encoder $f_{\theta}(\cdot)$ and a projection head $q_{\psi}(\cdot)$, which are trained with the contrastive objective $\mathcal{L}_{cl}$. Finally, we discard the projection head and fine-tune the encoder with an untrained decoder $g_{\phi}(\cdot)$ to predict the future via optimizing $\mathcal{L}_{pred}$.

\subsubsection{Joint learning.}
In the joint learning approach, we need to simultaneously perform forecasting and contrastive tasks. However, using the original input for forecasting and two augmented views for contrasting induces unnecessary overhead. Instead, we use the original input $\mathcal{G}$ not only to conduct the forecasting task but also as the first view ($\mathcal{G}^{\prime}=\mathcal{G}$), and only use augmentations to form the second view $\mathcal{G}^{\prime\prime}$ for contrastive learning.

Subsequently, an STG encoder $f_{\theta}(\cdot)$ is jointly trained with a projection head $q_{\psi}(\cdot)$ and an STG decoder $g_{\phi}(\cdot)$. In this case, the contrastive objective serves as an auxiliary regularization term during training and provides additional self-supervision signals for improving generalization \cite{you2020does}. The final form of the loss function is shown below, where $\lambda$ is a trade-off parameter.
\begin{equation}
    \mathcal{L} = \mathcal{L}_{pred} + \lambda\mathcal{L}_{cl}
    \label{eq4}
\end{equation}

\begin{figure}[!t]
  \centering
  \vspace{-1em}
  \includegraphics[width=0.95\linewidth]{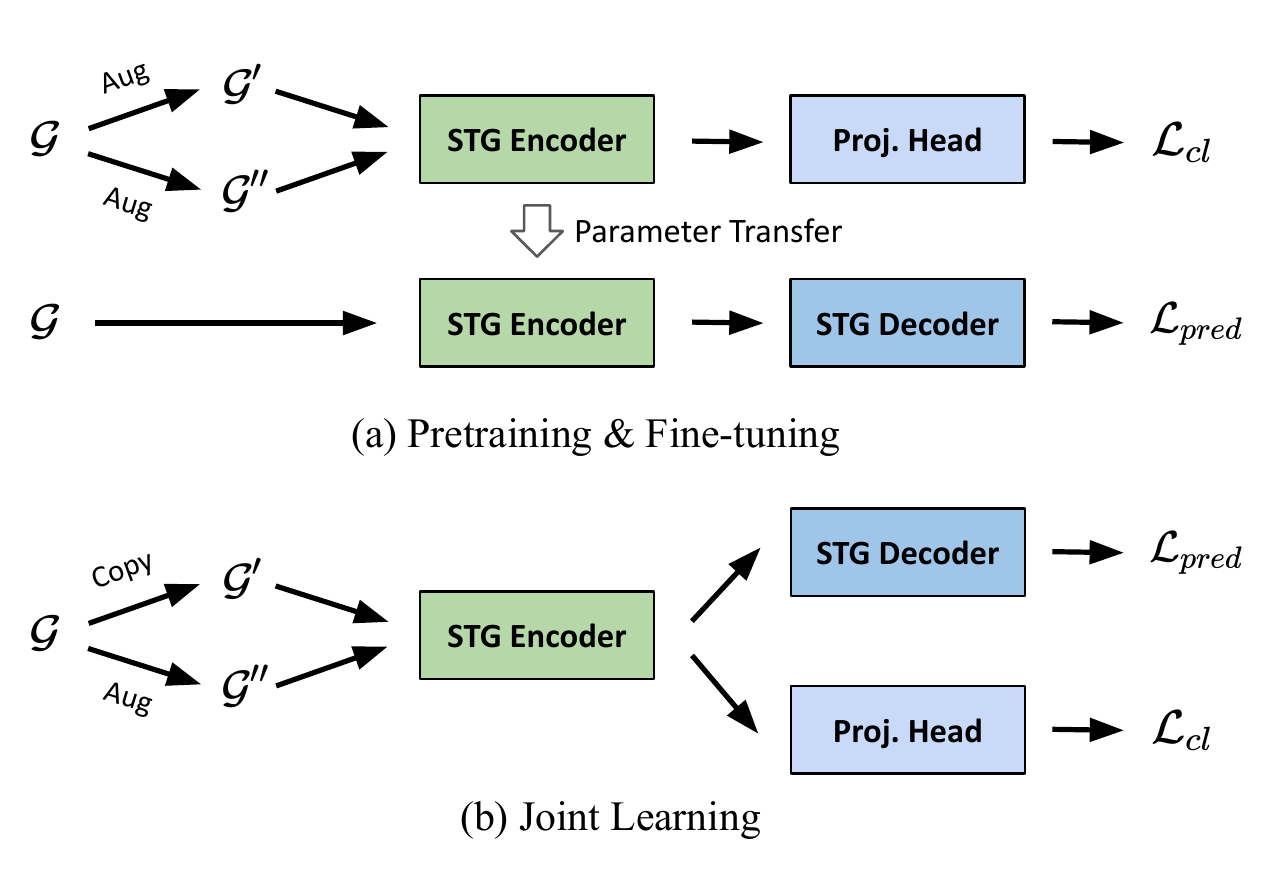}
  \vspace{-1em}
  \caption{Illustration of different training schemes.}
  \label{fig:scheme}
\end{figure}

\subsection{What \& How to Contrast (Q2)}
\label{sec:3.2}
As described in Section \ref{sec:2.3}, the high-level representation extracted from an STG encoder is $\mathbf{H} \in \mathbb{R}^{N \times D}$, which means that each node has its latent features. Hence, we propose two feasible designs of contrastive learning methods, which contrast on different objects, i.e., node- or graph-level representations. The rationales behind the two designs are different. For node-level contrast, it is more fine-grained and matches to the level of the forecasting tasks, i.e., generating predictions at each node. In graph-level contrast, it considers global knowledge of the whole graph, which may aid each node for learning a more useful representation.

\subsubsection{Node-level contrast.}
In node-level method, we directly take $\mathbf{H}$ and use a projection network $q_{\psi}(\cdot)$ to map it to another latent space $\mathbf{Z} \in \mathbb{R}^{N \times D}$. Suppose we have a batch of $M$ STG with $N$ nodes, given an anchor (a node representation), its positive comes from its augmented view. To perform full spatio-temporal contrast, its candidate negatives include all the other nodes in this time step and all the nodes in other time steps in this batch, resulting in $MN$ negatives in total. When computing pair-wise cosine similarity, this method incurs high computational costs and memory usage (the computational complexity is $O(M^{2}N^{2})$), which may become totally unaffordable in large graphs. 

\begin{figure}[!h]
% \vspace{-0.5em}
  \centering
  \includegraphics[width=\linewidth]{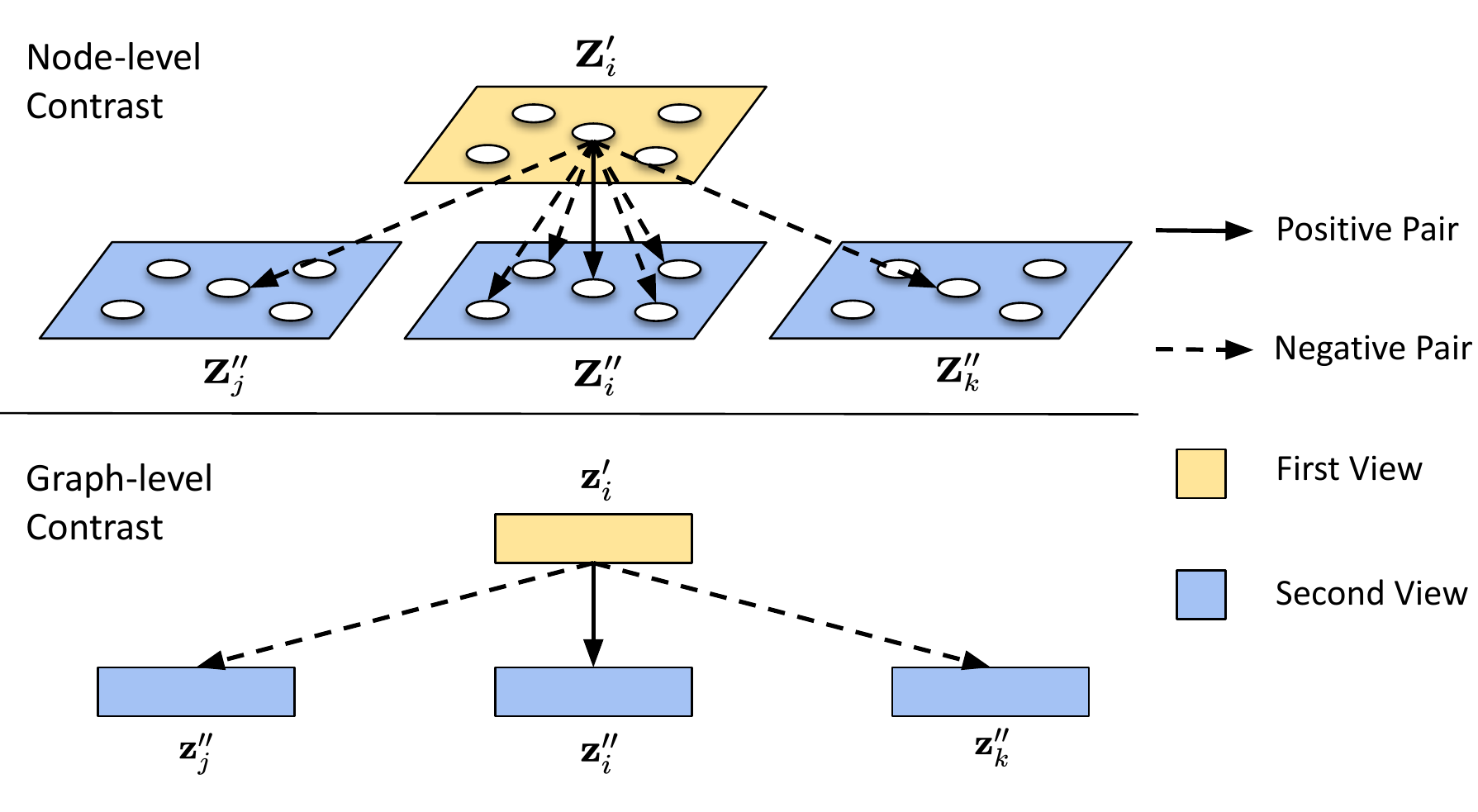}
  \vspace{-1.5em}
  \caption{Sketch of node- and graph-level contrast. The index $i$ denotes the anchor and indices $j, k$ denote two arbitrary STG instances in this batch.}
  \label{fig:level}
\end{figure}

Inspired by existing arts \cite{wu2019graph, bai2020adaptive} that capture spatio-temporal dependencies separately, we address this issue by factorizing the full spatio-temporal contrast along the spatial and temporal dimensions. The negatives now come from the other nodes at this time step and the same node at the other time steps (see Figure \ref{fig:level}). In this way, we only have $M + N$ negatives and the complexity decreases to $O(M^{2} + N^{2})$, leading to better efficiency and memory utility.

\subsubsection{Graph-level contrast.}
In this approach, we first utilize a summation function as a readout function to obtain graph-level representation $\mathbf{s} \in \mathbb{R}^{D}$. This representation can be seen as a summary of a whole STG input. Then we apply a projection head $q_{\psi}(\cdot)$ to map it to the latent space $\mathbf{z} \in \mathbb{R}^{D}$. Given an anchor (a graph representation), its positive comes from its augmented version. Its potential negatives are the representations from the other graphs within this batch (see Figure \ref{fig:level}). By contrasting at this level, a model is encouraged to distinguish different inputs' spatio-temporal patterns.

% we propose a novel framework STGCL that enhances STG forecasting with contrastive learning and makes full use of the unique characteristics of STG data. Our goal is to encourage the spatio-temporal summaries obtained from the encoder to be invariant to perturbations and to be discriminative to distinguish different samples' spatio-temporal patterns. These will help to improve performance and strengthen the model's robustness. The pipeline of STGCL is depicted as follows (also see Figure \ref{fig:framework}).
% Most contrastive methods construct two augmented views, we only generate one augmented view. The reason is that if we generate two views, then we need to forward the model three times, which will cause a long training time.
% In most contrastive methods, they have to construct two augmented views based on original inputs. However, it will induce much computational overhead since we need to forward the model three times. Thus, ...

\subsection{Data Augmentation (Q3)}
\label{sec:3.3}
% Several augmentation methods on graphs have been proposed in \cite{you2020graph, zeng2021contrastive}, such as subgraph sampling. However, they are originally designed for conventional graphs, which is not the case for STG. For example, they do not consider the temporal correlations.
% We assess each of the proposed methods and find that their performance only have minor differences, which indicates that the models are not sensitive to the semantic/assumption of augmentations.
Data augmentation is a necessary component of the contrastive learning techniques. It helps to build semantically similar pairs and assists the model to learn invariant representations under different types and levels of perturbations \cite{chen2020simple}. However, so far data augmentations have been less explored for STG. For example, the intrinsic properties of STG data (especially temporal dependencies) are not well considered in current graph augmentation methods \cite{you2020graph, zeng2021contrastive}.

In this work, we modify two popular graph augmentations (edge perturbation \cite{you2020graph} and attribute masking \cite{you2020graph}) and propose two new approaches specifically designed for STG. The four methods perturb data in terms of graph structure, time domain, and frequency domain, thus making the learned representation less sensitive to changes of graph structure or signals. We denote $\mathbf{X}^{(t-S):t} \in \mathbb{R}^{S \times N}$ in this section, because only the target attribute (e.g., traffic speed) is modified. The details of each method are given as follows.
% multivariate time series?

\subsubsection{Edge masking.} 
Edge perturbation \cite{you2020graph} suggests either adding or deleting a certain ratio of edges to/from an unweighted graph. However, it is an awkward fit for the weighted adjacency matrix used in STG forecasting, since assigning proper weights to the added edges is difficult. Therefore, we make a revision by masking (deleting) entries of the adjacency matrix to perturb the graph structure. Each entry of the augmented matrix $\mathbf{A}^{\prime}$ is given by:
\begin{equation}
    \mathbf{A}^{\prime}_{ij} = 
        \begin{cases}
            \mathbf{A}_{ij}, & \text{if} \ \mathbf{M}_{ij} \geqslant r_{em} \\
            0, & \text{otherwise}
        \end{cases}
\end{equation}
where $\mathbf{M} \sim U(0, 1)$ is a random matrix and $r_{em}$ is tunable. We share the augmented matrix across the inputs within a batch for efficiency. This method is applicable for both predefined and adaptive adjacency matrix \cite{wu2019graph}.
% because creating an augmented adjacency matrix for each sample will cause a large overhead when computing graph convolution in the model.
% to be more aware of the local structures of a node

\subsubsection{Input masking.}
% As mentioned before, there are often some missing values in STG data. To strengthen the model robustness to this factor, we simulate this process by masking entries of the original input feature matrix. 
We adjust the attribute masking \cite{you2020graph} method by restricting the mask only on the target attribute. The goal is to strengthen the model's robustness to the factor of missing values, which is a general case in STG data. Hence, each entry of the augmented feature matrix $\mathbf{P}^{(t-S):t}$ is generated by:
\begin{equation}
    \mathbf{P}^{(t-S):t}_{ij} = 
        \begin{cases}
            \mathbf{X}^{(t-S):t}_{ij}, & \text{if} \ \mathbf{M}_{ij} \geqslant r_{im} \\
            -1, & \text{otherwise}
        \end{cases}
\end{equation}
where $\mathbf{M} \sim U(0, 1)$ is a random matrix and $r_{im}$ is tunable.
% Masked entries are replaced by the mean, since the input feature is preprocessed by z-score normalization. Can add this sentence in exp section later

\subsubsection{Temporal shifting.} 
\begin{figure}[!t]
  \centering
  \includegraphics[width=\linewidth]{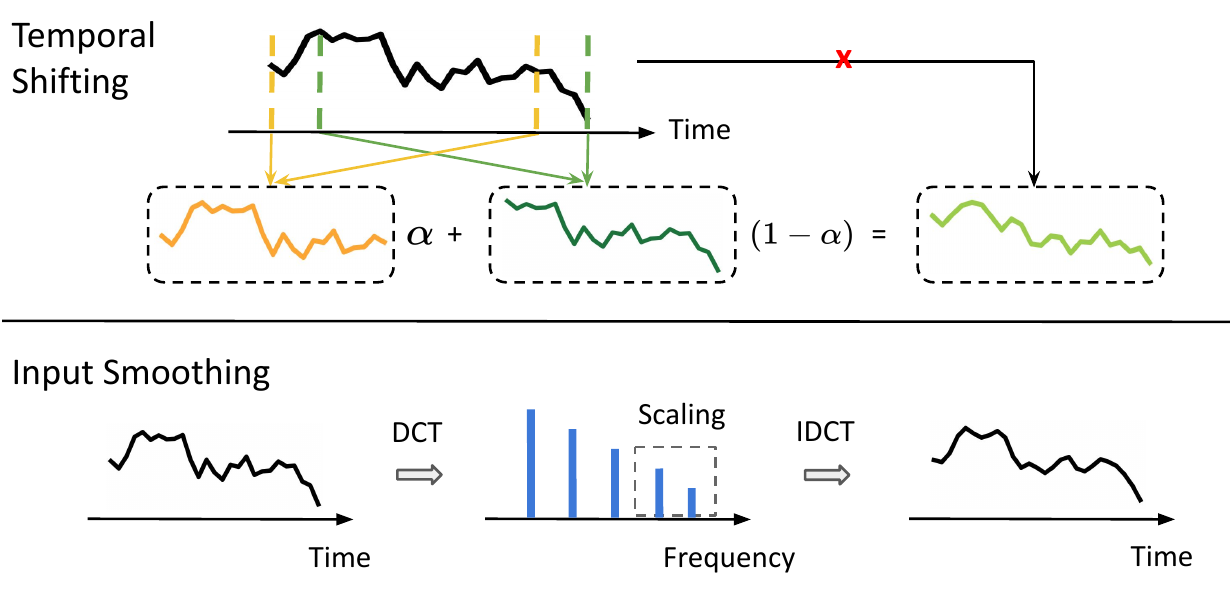}
  \caption{Sketch of temporal shifting and input smoothing.}
  \label{fig:aug-sketch}
%   \vspace{-1em}
\end{figure}

STG data derive from nature and evolve over time continuously. However, they can only be recorded by sensors in a discrete manner, e.g., a reading every five minutes. Motivated by this, we shift the data along the time axis to exploit the intermediate status between two consecutive time steps (see Figure \ref{fig:aug-sketch}). We implement this idea by linearly interpolating between consecutive inputs. Formally,
\begin{equation}
    \mathbf{P}^{(t-S):t} = \alpha\mathbf{X}^{(t-S):t} + (1 - \alpha)\mathbf{X}^{(t-S+1):(t+1)}
\end{equation}
where $\alpha$ is generated within the distribution $U(r_{ts}, 1)$ every epoch and $r_{ts}$ is adjustable. This method is input-specific, which means different inputs have their unique $\alpha$. Meanwhile, our operation can be related to the mixup augmentation \cite{zhang2017mixup}. The major difference is that we conduct weighted averaging between two successive time steps to ensure interpolation accuracy.

\subsubsection{Input smoothing.}
To reduce the impact of data noise in STG, this method smooths the inputs by scaling high-frequency entries in the frequency domain (see Figure \ref{fig:aug-sketch}). Specifically, we first concatenate histories with future values (both are available during training) to enlarge the length of the time series to $L = S + T$, and obtain $\mathbf{X}^{(t-S):(t+T)} \in \mathbb{R}^{L \times N}$. Then we apply Discrete Cosine Transform (DCT) to convert the sequence of each node from the time domain to the frequency domain. We keep the low frequency $E_{is}$ entries unchanged and scale the high frequency $L - E_{is}$ entries by the following steps: 1) We generate a random matrix $\mathbf{M} \in \mathbb{R}^{(L - E_{is}) \times N}$, which satisfies $\mathbf{M} \sim U(r_{is}, 1)$ and $r_{is}$ is adjustable. 2) We leverage normalized adjacency matrix $\mathbf{\tilde{A}}$ to smooth the generated matrix by $\mathbf{M} = \mathbf{M}\mathbf{\tilde{A}}^{2}$. The intuition is that neighboring sensors should have similar scaling ranges and multiplying two steps of $\mathbf{\tilde{A}}$ should be sufficient to smooth the data. This step can be omitted when the adjacency matrix is not available. 3) We element-wise multiply the random numbers with the original $L - E_{is}$ entries. Finally, we use Inverse DCT (IDCT) to convert data back to the time domain.
% add DCT and IDCT formula here or not
% \paragraph{Input noising} This method injects noise in the time domain. We first generate Gaussian noise $\mathbf{M}_{in} \in \mathbb{R}^{S \times N}$, where each entry of $\mathbf{M}_{in}$ is generated from a $\mathcal{N}(0, \sigma_{in}^{2})$ distribution and $\sigma_{in}$ is tunable. Then we add this Gaussian noise to the input feature matrix.

\subsection{Negative Filtering (Q4)}
\label{sec:3.4}
In most contrastive learning methods, all the other objects within a batch are seen as negatives for a given anchor. This ignores the fact that some of these objects may be semantically similar to the anchor, and are inappropriate to be used as negatives. There are some solutions suggesting using instance’s label to assign positive and negative pairs \cite{khosla2020supervised, han2020self}. However, STG forecasting is a typical regression task without semantic labels. Thus, we propose to filter out unsuitable negatives based on the unique properties of STG data. Denoting $\chi_i$ as the set of acceptable negatives for the $i$th object after filtering, we extend the contrastive loss in Eq. \ref{eq1} to:
\begin{equation}
    \mathcal{L}_{cl} = \frac{1}{M} \sum_{i=1}^{M} - \log \frac{\exp({\rm sim}(\mathbf{z}_{i}^{\prime}, \mathbf{z}_{i}^{\prime\prime}) / \tau)}{\sum_{j \in \chi_i} \exp({\rm sim}(\mathbf{z}_{i}^{\prime}, \mathbf{z}_{j}^{\prime\prime}) / \tau)}
    \label{eq2}
\end{equation}

\subsubsection{Spatial negative filtering}
In the spatial dimension, we utilize the information from the predefined adjacency matrix. Specifically, the first-order neighbors of each node are excluded during contrastive loss computation. Note that this part is only applicable for node-level contrast.

\subsubsection{Temporal negative filtering}
\begin{figure}[!h]
  \centering
  \vspace{-1em}
  \includegraphics[width=0.9\linewidth]{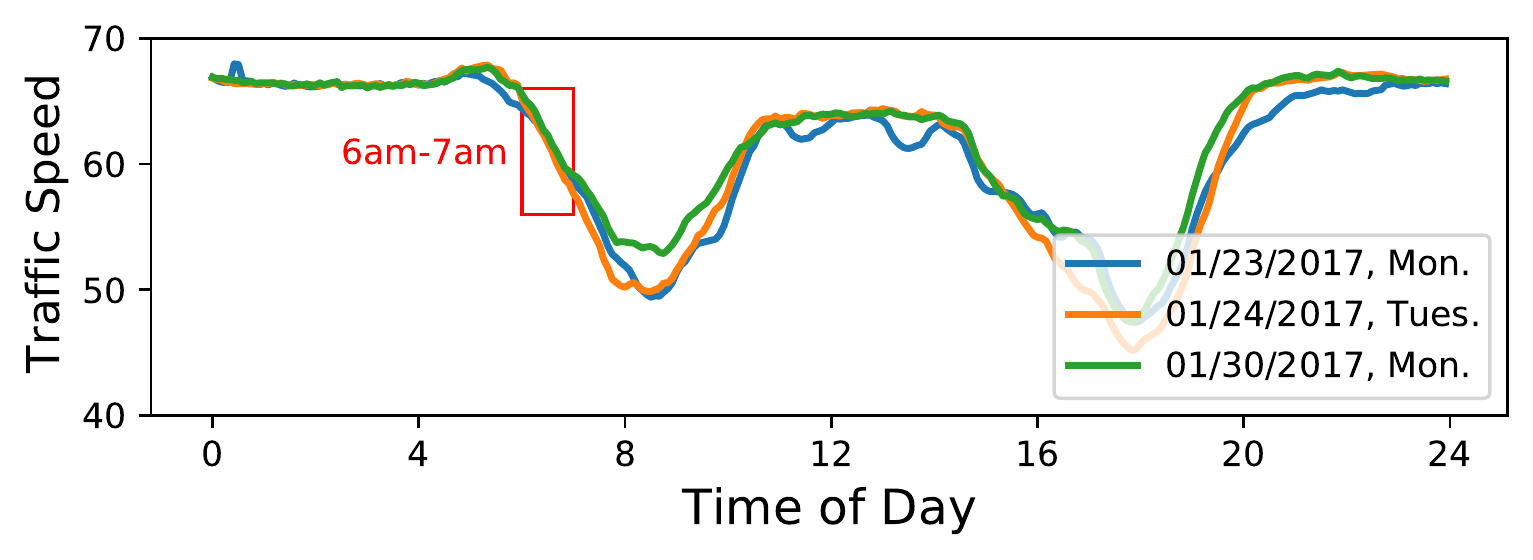}
  \vspace{-1em}
  \caption{Example of temporal correlations in STG data.}
  %Each point in this figure is the average traffic speed of all nodes at that moment.}
  \label{fig:periodic}
\end{figure}

Unlike the former method, temporal negative filtering can be applied to both levels. Here, we propose to filter negatives by utilizing the ubiquitous temporal properties of STG data -- closeness and periodicity \cite{zhang2017deep}. To facilitate understanding, Figure \ref{fig:periodic} depicts a screenshot of the traffic speed in a time period on the BAY dataset \cite{li2018diffusion}. We can observe that the pattern from 6 am to 7 am on Monday is similar to that day from 7 am to 8 am (closeness), and also similar to the same time on Tuesday (daily periodic) and next Monday (weekly periodic). Hence, temporally close inputs (regardless of the day) are likely to have similar spatio-temporal representations in the latent space. If we form negative pairs by using these semantically similar inputs (i.e., hard negatives) and push apart their representations, it may break the semantic structure and worsen performance.

Therefore, we devise a rule-based strategy that leverages the attribute ``time of day'' in the input to filter out the hardest negatives (i.e., the most similar inputs in semantics) per anchor. In this forecasting task, due to the lack of semantic labels, we define the hardest negatives using a controllable threshold $r_f$. Specifically, denoting the starting ``time of day'' of an anchor as $t_{i}$, we obtain the acceptable negatives within a batch by letting each input's starting ``time of day'' $t$ satisfy the requirement of $|t - t_i | > r_f$. If $r_f$ is set to zero, all the other objects in a batch are used to form the negative pairs with the anchor. We empirically find that the choice of $r_f$ is of great importance to the quality of learned representations. For instance, using a very large $r_f$ will significantly reduce the number of negatives, which may impair the contrastive task. For more details, please see Section \ref{sec:4.4.2}.

% Increasing $r_f$ expands the filtering range, excludes more hard negative samples, and makes the contrastive task easier, but the discriminative representations that can be learned become limited.
% More detailed analysis and interpretation of how the negative filtering helps contrastive learning will be in Section \ref{sec:exp-loss}.
% Since less negatives are incorporated, this filtering operations enjoy less computational overhead.
% Several intuitive solutions in the vision domain suggest using label information to assign positive and negative pairs \cite{khosla2020supervised, han2020self}. However, STG forecasting is a typical regression task without semantic labels. It is difficult to judge whether two samples can be regarded as a positive pair. Here we argue that it is relatively safe to say that the hardest negative samples (the most similar in semantics) are not suitable for forming negative pairs with the anchor.

\subsection{Implementation}
\label{sec:3.5}
\begin{figure}[!t]
  \centering
  \includegraphics[width=\linewidth]{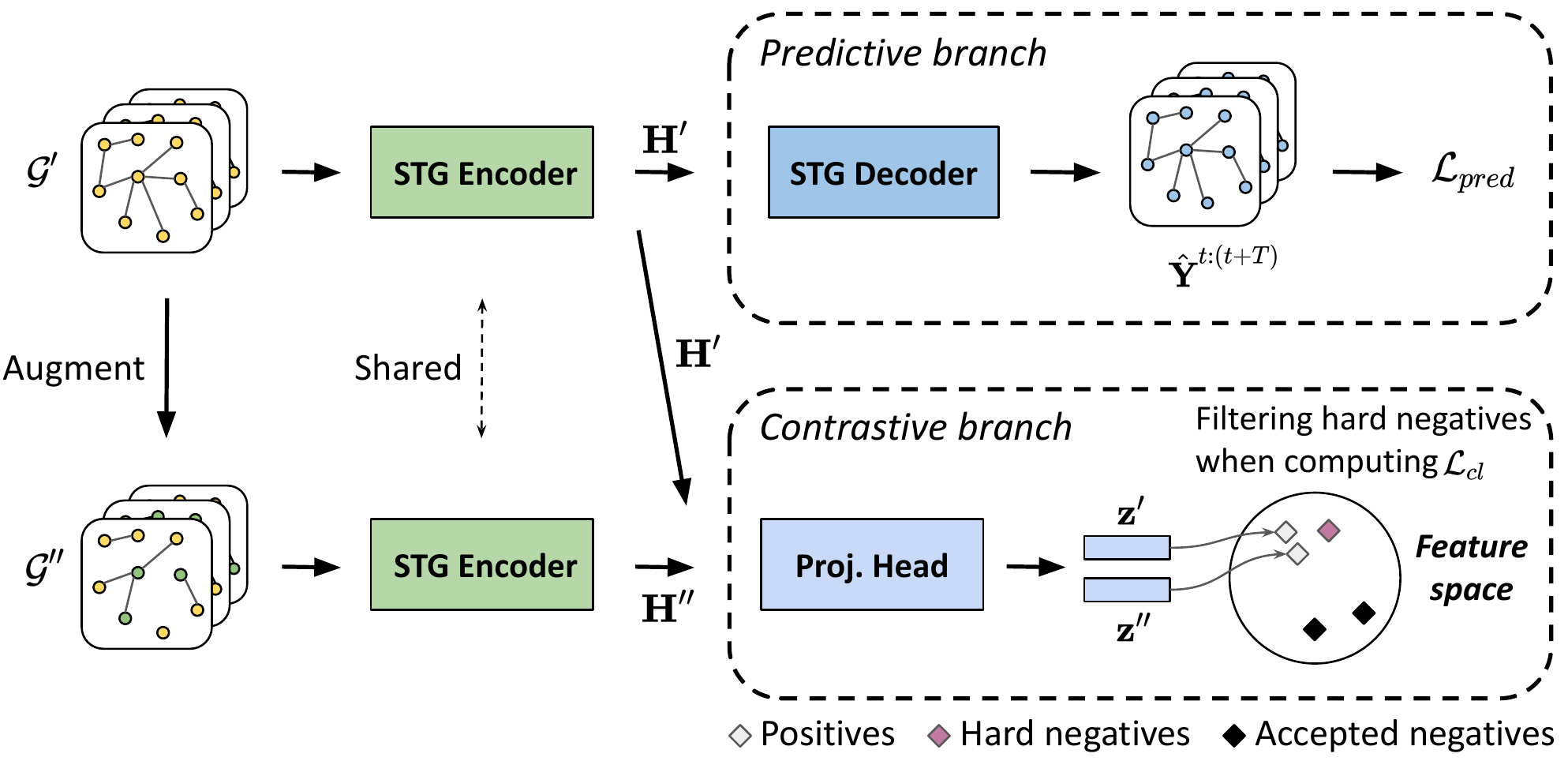}
  \vspace{-1em}
  \caption{Sample implementation of STGCL. The missing edges and green nodes in $\mathcal{G}^{\prime\prime}$ indicate the perturbations on graph structure and signals, respectively.}
  \label{fig:framework}
  \vspace{-0.5em}
\end{figure}

To ease understanding, here, we give a sample implementation to link all the techniques introduced above. Figure \ref{fig:framework} shows the implementation of choosing joint learning scheme and contrasting at graph level. Specifically, we first utilize an STG encoder to map both the original input and the augmented input to high-level representations $\mathbf{H}^{\prime}, \mathbf{H}^{\prime\prime}$. Then, the representations flow into the following two branches.
\begin{itemize}[leftmargin=*]
    \item A \textit{predictive branch} that feeds the representation $\mathbf{H}^{\prime}$ through an STG decoder to forecast the future steps. The predictions of the decoder $\hat{\mathbf{Y}}^{t:(t+T)}$ are used to compute the forecasting loss with the ground truth.
    \item A \textit{contrastive branch} that takes both $\mathbf{H}^{\prime}$ and $\mathbf{H}^{\prime\prime}$ as inputs to conduct the auxiliary contrastive task. Specifically, we utilize a readout function to obtain the spatio-temporal summaries $\mathbf{s}^{\prime}, \mathbf{s}^{\prime\prime} \in \mathbb{R}^{D}$ of the input. We further map the summaries to the latent space $\mathbf{z}^{\prime}, \mathbf{z}^{\prime\prime} \in \mathbb{R}^{D}$ by a projection head. The applied projection network has two linear layers, where the first layer is followed by batch normalization and rectified linear units. Afterwards, a contrastive loss is used to maximize the similarities between $\mathbf{z}^{\prime}$ and $\mathbf{z}^{\prime\prime}$, and minimize the similarities between $\mathbf{z}^{\prime}$ and other inputs' augmented view. Lastly, we use negative filtering to avoid forming negative pairs between the most semantically similar inputs.
\end{itemize}

In addition, Algorithm \ref{alg:algorithm} summarizes the training algorithm of the implementation. Line 6 is the code for predictive branch. Line 8-10 include the code for contrastive branch. Note that the data augmentation module $\operatorname{aug}()$ comprises the four proposed augmentation methods. The methods can be applied either solely or jointly.

% \vspace{-1em}
\begin{algorithm}[!t]
\LinesNumbered 
\KwIn{Training data $\mathcal{G}$, labels $\mathbf{Y}$, constant $r_f$, $\tau$, $\lambda$, \\
$\quad\quad\quad$ STG encoder $f_{\theta}(\cdot)$ and decoder $g_{\phi}(\cdot)$, \\ 
$\quad\quad\quad$ projection head $q_{\psi}(\cdot)$, \\
$\quad\quad\quad$ augmentation module $\operatorname{aug}()$, \\
$\quad\quad\quad$ readout function $\operatorname{readout}()$, \\
$\quad\quad\quad$ negative filtering operation $\operatorname{filter}()$.}
\KwOut{$f_{\theta}(\cdot)$ and $g_{\phi}(\cdot)$.} 
{ %[1] enables line numbers
    \For{sampled batch $\{\mathcal{G}_{i}\}_{i=1}^{M} \in \mathcal{G}$}{
        \For{$i = 1$ \textbf{to} $M$}{
            $\mathcal{G}_{i}^{\prime} =\mathcal{G}_{i}$\;
            $\mathcal{G}_{i}^{\prime\prime} = \operatorname{aug}(\mathcal{G}_{i})$\;
            $\mathbf{H}_{i}^{\prime}, \mathbf{H}_{i}^{\prime\prime} = f_{\theta}(\mathcal{G}_{i}^{\prime}), f_{\theta}(\mathcal{G}_{i}^{\prime\prime})$\;
            \# predictive branch\;
            $\hat{\mathbf{Y}}_{i} = g_{\phi}(\mathbf{H}_{i}^{\prime})$\;
            \# contrastive branch\;
            $\mathbf{s}_{i}^{\prime}, \mathbf{s}_{i}^{\prime\prime} = \operatorname{readout}(\mathbf{H}_{i}^{\prime}), \operatorname{readout}(\mathbf{H}_{i}^{\prime\prime})$\;
            $\mathbf{z}_{i}^{\prime}, \mathbf{z}_{i}^{\prime\prime} = q_{\psi}(\mathbf{s}_{i}^{\prime}), q_{\psi}(\mathbf{s}_{i}^{\prime\prime})$\;
            $\chi_i \leftarrow \operatorname{filter}(\mathcal{G}_{i}, r_f)$\;
        }
    $\mathcal{L}_{pred} = \frac{1}{M} \sum_{i=1}^{M} |\hat{\mathbf{Y}}_{i} - \mathbf{Y}_{i}|$\;
    $\mathcal{L}_{cl} = \frac{1}{M} \sum_{i=1}^{M} - \log \frac{\exp({\rm sim}(\mathbf{z}_{i}^{\prime}, \mathbf{z}_{i}^{\prime\prime}) / \tau)}{\sum_{j \in \chi_i} \exp({\rm sim}(\mathbf{z}_{i}^{\prime}, \mathbf{z}_{j}^{\prime\prime}) / \tau)}$\;
    update $f_{\theta}(\cdot)$, $g_{\phi}(\cdot)$, and $q_{\psi}(\cdot)$ to minimize $\mathcal{L}_{pred} + \lambda\mathcal{L}_{cl}$\;
    }
    \textbf{return} $f_{\theta}(\cdot)$ and  $g_{\phi}(\cdot)$\;
}
% \end{algorithmic}
\caption{Training algorithm of the implementation.}
\label{alg:algorithm}
\end{algorithm}

\section{Experiments}
In this section, we broadly evaluate, investigate, and interpret the benefits of integrating contrastive learning into STG forecasting. We first detail the experimental setup in Section \ref{sec:4.1}. Then, we present the empirical results of using different training schemes and contrastive learning methods in Section \ref{sec:4.2} and \ref{sec:4.3}, respectively. Finally, we provide an ablation study on the effects of data augmentation and negative filtering in Section \ref{sec:4.4}.

\subsection{Experimental Setup}
\label{sec:4.1}
\subsubsection{Datasets}
% BAY dataset is from \cite{li2018diffusion}
We assess STGCL on two popular traffic benchmarks from literature: PEMS-04 and PEMS-08 \cite{guo2019attention}. To have a fair comparison to prior works, an additional evaluation dataset is included in Table \ref{tab:perf-appendix}. There are three kinds of traffic measurements contained in the raw data, including traffic flow, average speed, and average occupancy. These traffic readings are aggregated into 5-minute windows, resulting in 288 data points per day. Following common practice \cite{guo2019attention}, we use traffic flow as the target attribute and use the 12-step historical data to predict the next 12 steps. Z-score normalization is applied to the input data for fast training. The adjacency matrix is constructed by road-network distance with the thresholded Gaussian kernel method \cite{shuman2013emerging}. The datasets are divided into three parts for training, validation, and testing with a ratio of 6:2:2. More details are provided in Table \ref{tab:dataset} and Appendix \ref{sec:dataset-appendix}.
% The datasets are collected by the Caltrans Performance Measurement System \cite{chen2001freeway}.

\begin{table}[!h]
\vspace{-0.5em}
\centering
\small
\tabcolsep=2.5mm
  \caption{Dataset statistics.}
  \vspace{-0.5em}
  \label{tab:dataset}
  \begin{tabular}{l|cccc}
    \shline
     Datasets & \#Nodes & \#Edges & \#Instances & Interval \\
    \hline
    %  BAY & 325 & 2,369 & 52,116 & 5 min \\
     PEMS-04 & 307 & 209 & 16,992 & 5 min \\
     PEMS-08 & 170 & 137 & 17,856 & 5 min \\
    \shline
  \end{tabular}
  \vspace{-0.5em}
\end{table}

\subsubsection{Base Models}
According to the performance comparison presented at LibCity\footnote{https://libcity.ai/} and \cite{li2021dynamic}, we select four models that have the top performance. They belong to the following two classes: CNN-based models (GWN, MTGNN) and RNN-based models (DCRNN, AGCRN).
\begin{itemize}[leftmargin=*]
    \item GWN: Graph WaveNet combines adaptive adjacency matrix with graph convolution and uses dilated casual convolution \cite{wu2019graph}.
    \item MTGNN: This approach proposes a graph learning module and applies graph convolution with mix-hop propagation and dilated inception layer \cite{wu2020connecting}.
    \item DCRNN: Diffusion Convolution Recurrent Neural Network, which integrates diffusion convolution into RNNs  \cite{li2018diffusion}.
    \item AGCRN: This approach develops two adaptive modules to enhance graph convolution and combines them into RNNs \cite{bai2020adaptive}.
\end{itemize}

\subsubsection{Experiment Details}
We use PyTorch 1.8 to implement all the base models and our method. We basically employ the default settings of base models from their source code, such as model configurations, batch size, optimizer, learning rate, and gradient clipping. The model-related settings of STGCL follow the same as in base models’ implementation, except that we don’t apply weight decay, as this will introduce another term in the loss function. For STGCL's specific settings, the temperature $\tau$ is set to 0.1 and the usage of augmentations and $r_f$ are described in Section \ref{sec:4.4}. All experiments are repeated five times with different seeds. We adopt three common metrics in forecasting tasks to evaluate the model performance, including mean absolute error (MAE), root mean squared error (RMSE), and mean absolute percentage error (MAPE). See more details about the settings in Appendix \ref{sec:base-appendix} and \ref{sec:stgcl-appendix}.

\begin{table}[!t]
\tabcolsep=3.5mm
\caption{Test MAE results of the average values over all predicted time steps when integrating two contrastive tasks (node and graph) into two base models (GWN and AGCRN) through two training schemes: pretraining \& fine-tuning (P\&F) and joint learning (JL). We highlight the best performance in bold font.}
\label{tab:scheme}
\begin{tabular}{l|c|c}
    \shline
    Methods & PEMS-04 & PEMS-08 \\ \hline
    GWN & 19.33$\pm$0.11 & 14.78$\pm$0.03 \\
    w/ P\&F-node & 20.22$\pm$0.22 & 15.37$\pm$0.08 \\
    w/ P\&F-graph & 20.67$\pm$0.13 & 15.86$\pm$0.15 \\
    w/ JL-node & 18.89$\pm$0.05 & 14.63$\pm$0.07 \\
    w/ JL-graph & \textbf{18.88$\pm$0.04} & \textbf{14.61$\pm$0.03} \\ \hline\hline
    AGCRN & 19.39$\pm$0.03 & 15.79$\pm$0.06 \\
    w/ P\&F-node & 19.70$\pm$0.07 & 17.21$\pm$0.14 \\
    w/ P\&F-graph & 20.39$\pm$0.10 & 17.92$\pm$0.10 \\
    w/ JL-node & 19.32$\pm$0.06 & 15.78$\pm$0.09 \\
    w/ JL-graph & \textbf{19.13$\pm$0.05} & \textbf{15.62$\pm$0.07} \\
    \shline
\end{tabular}
\vspace{-0.5em}
\end{table}

\begin{table*}[t]
\centering
\tabcolsep=2.5mm
\caption{Test MAE results of three specific horizons when incorporating two contrastive tasks into four state-of-the-art models through joint learning (JL). $\Delta$ denotes the sum of improvements against the base models for the three horizons. min: minutes.}
\label{tab:perf}
\vspace{-1em}
\begin{tabular}{l|cccc|cccc}
\shline
\multirow{2}{*}{Methods} & \multicolumn{4}{c|}{PEMS-04} & \multicolumn{4}{c}{PEMS-08} \\ \cline{2-9}
& 15 min & 30 min & 60 min & $\Delta$ & 15 min & 30 min & 60 min & $\Delta$ \\ \hline

GWN & 18.20$\pm$0.09 & 19.32$\pm$0.13 & 21.10$\pm$0.18 & -- & 13.80$\pm$0.05 & 14.75$\pm$0.04 & 16.39$\pm$0.09 & -- \\

w/ JL-node & 17.97$\pm$0.05 & 18.90$\pm$0.07 & \textbf{20.38$\pm$0.07} & -1.37 & 13.67$\pm$0.06 & 14.63$\pm$0.08 & 16.14$\pm$0.13 & -0.50 \\

w/ JL-graph & \textbf{17.93$\pm$0.04} & \textbf{18.87$\pm$0.04} & 20.40$\pm$0.09 & \textbf{-1.42} & \textbf{13.67$\pm$0.04} & \textbf{14.61$\pm$0.03} & \textbf{16.09$\pm$0.05} & \textbf{-0.57} \\ \hline\hline

MTGNN & 18.32$\pm$0.05 & 19.10$\pm$0.05 & 20.39$\pm$0.09 & -- & 14.36$\pm$0.06 & 15.34$\pm$0.10 & 16.91$\pm$0.16 & -- \\

w/ JL-node & 18.03$\pm$0.02 & 18.79$\pm$0.06 & 19.94$\pm$0.03 & -1.05 & 14.05$\pm$0.05 & 14.94$\pm$0.04 & 16.38$\pm$0.09 & -1.24 \\

w/ JL-graph & \textbf{17.99$\pm$0.03} & \textbf{18.72$\pm$0.05} & \textbf{19.88$\pm$0.07} & \textbf{-1.22} & \textbf{14.04$\pm$0.05} & \textbf{14.90$\pm$0.05} & \textbf{16.23$\pm$0.08} & \textbf{-1.44} \\ \hline\hline

DCRNN & 19.99$\pm$0.11 & 22.40$\pm$0.19 & 27.15$\pm$0.35 & -- & 15.23$\pm$0.15 & 16.98$\pm$0.25 & 20.27$\pm$0.41 & -- \\

w/ JL-node & 19.94$\pm$0.08 & 22.38$\pm$0.14 & 27.15$\pm$0.26 & -0.07 & \textbf{15.15$\pm$0.05} & \textbf{16.85$\pm$0.11} & \textbf{20.02$\pm$0.23} & \textbf{-0.46} \\

w/ JL-graph & \textbf{19.82$\pm$0.08} & \textbf{22.07$\pm$0.12} & \textbf{26.51$\pm$0.21} & \textbf{-1.14} & 15.19$\pm$0.11 & 16.89$\pm$0.19 & 20.09$\pm$0.34 & -0.31 \\ \hline\hline

AGCRN & 18.53$\pm$0.03 & 19.43$\pm$0.06 & 20.72$\pm$0.03 & -- & 14.58$\pm$0.07 & 15.71$\pm$0.07 & 17.82$\pm$0.11 & -- \\

w/ JL-node & 18.46$\pm$0.04 & 19.37$\pm$0.07 & 20.69$\pm$0.11 & -0.16 & 14.55$\pm$0.06 & 15.69$\pm$0.10 & 17.82$\pm$0.14 & -0.05 \\

w/ JL-graph & \textbf{18.31$\pm$0.04} & \textbf{19.17$\pm$0.06} & \textbf{20.39$\pm$0.03} & \textbf{-0.81} & \textbf{14.51$\pm$0.05} & \textbf{15.56$\pm$0.06} & \textbf{17.51$\pm$0.10} & \textbf{-0.53} \\ \shline
\end{tabular}
\end{table*}

\subsection{Training Schemes Evaluation (Q1)}
\label{sec:4.2}
\begin{figure}[!t]
  \centering
  \includegraphics[width=\linewidth]{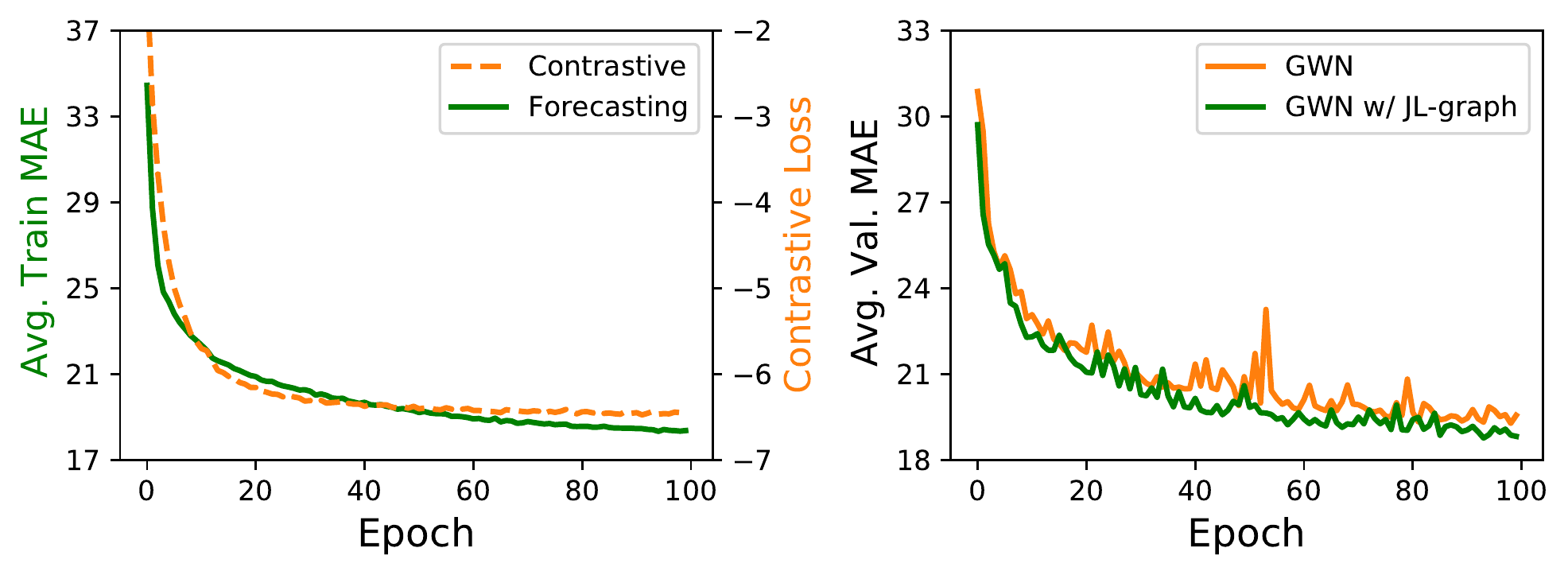}
  \caption{The left part shows the training curves of forecasting loss and contrastive loss by using GWN w/ JL-graph on PEMS-04. The right part shows the validation loss of GWN w/ JL-graph against pure GWN on PEMS-04.}
  \label{fig:loss}
  \vspace{-1em}
\end{figure}

We first examine the two training schemes mentioned in Q1, i.e., pretraining \& fine-tuning and joint learning, to integrate contrastive learning with STG forecasting. Table \ref{tab:scheme} summarizes the experimental results on two datasets by using two base models (GWN and AGCRN) and two contrastive learning tasks (node and graph level). The results demonstrate the effectiveness of collectively performing the forecasting and contrastive learning tasks, no matter what the models and contrastive objects are applied. This fact indicates that joint learning is the preferable way to provide additional self-supervision signals. We also plot the learning curves of the two losses ($\mathcal{L}_{cl}$ and $\mathcal{L}_{pred}$) in Figure \ref{fig:loss} (left hand side) to verify the learning process of joint learning. The reason why the contrastive loss is less than zero is that the denominator of the contrastive loss does not contain the numerator, thus their ratio can be greater than one. While on the right hand side of the figure, we show the trends of the validation loss. It is clear that GWN w/ JL-graph has better performance and is more stable than the pure GWN model.

Then, we investigate \emph{whether pretraining by contrastive learning can really benefit STG forecasting}. Specifically, we use a smaller learning rate during fine-tuning, i.e., 10\% of the learning rate applied in the base model, with the expectation to retain the information that has learned during pretraining. From Table \ref{tab:scheme}, it can be seen that pretraining \& fine-tuning performs worse than the base model, revealing the inefficacy of pretraining by contrastive learning. Going into deeper, according to a recent work \cite{wang2020understanding}, contrastive learning can be viewed as a direct optimization of two properties -- \emph{alignment} and \emph{uniformity}. The uniformity enforces the representations to be uniformly distributed on the unit hypersphere, thus facilitating different classes to be well clustered and linearly separable. This distribution provides more benefits to a classification problem, but may not capture useful information for the forecasting.
% while in regression task, similar node/graph still get clustered

Additionally, in the pretraining \& fine-tuning setting, we observe that the node-level contrast method outperforms the graph-level contrast. The results indicate that we have to ensure the consistency between the pretraining task and the downstream task, i.e., the prior knowledge extracted from the graph-level contrast cannot help the forecasting task that is performed on the node level. On the contrary, we find graph-level contrast surpasses node-level contrast in the joint learning setting, which will be detailed in Section \ref{sec:4.3}.
% We also provide a case study to explain the reasons and provide insights.

\subsection{Node \& Graph-Level Contrast (Q2)}
\label{sec:4.3}
We have demonstrated the power of the joint learning scheme in improving STG forecasting. In this section, we explore how the two levels of contrastive tasks behave on four state-of-the-art models. Table \ref{tab:perf} presents the MAE results on two datasets. See the complete table (including MAE, RMSE, MAPE) in Table \ref{tab:perf-appendix}.

Our results show the effectiveness of utilizing contrastive loss over all base models on all datasets. In particular, the graph-level method consistently outperforms the base models, while the node-level method achieves improvements in most cases. According to the student t-test, 54\%/83\% of the improvements are statistically significant at level 0.05 for node/graph level, respectively. These findings suggest that the proposed contrastive methods are model-agnostic, i.e., they can be plugged into both CNN-based and RNN-based architectures. In addition, the standard deviation of contrastive methods is generally smaller than that of the base model, indicating the additional stability provided by them.

% Note that particularly on the best performing models such as GWN and MTGNN, the improvements of graph-level contrast against base model are in many cases several times the standard deviation, especially on PEMS-04 and PEMS-08, suggesting that they are statistically significant.
% It is also worth mentioning that some of the improvements derived from STGCL are even larger than using an advanced model. For instance, GWN obtains an average MAE of 19.33 on PEMS-04, while the MAE is reduced to 18.88 after applying STGCL, better than MTGNN (MAE=19.09).

\begin{figure}[!t]
  \centering
  \includegraphics[width=\linewidth]{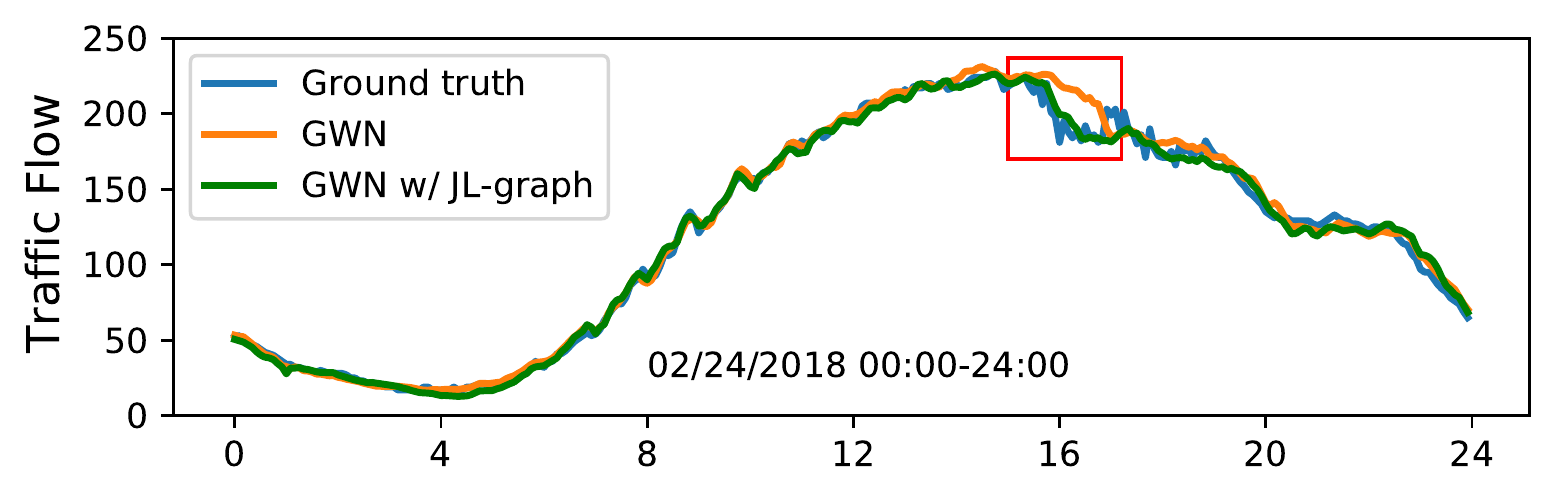}
  \vspace{-1.5em}
  \caption{Visualization of 60 minutes-ahead predictions on a snapshot of the PEMS-04 test set. x-axis: the time of day.}
  \label{fig:case}
%   \vspace{-1em}
\end{figure}

\begin{figure*}[!t]
     \centering
     \begin{subfigure}[b]{\textwidth}
         \centering
         \includegraphics[width=\textwidth]{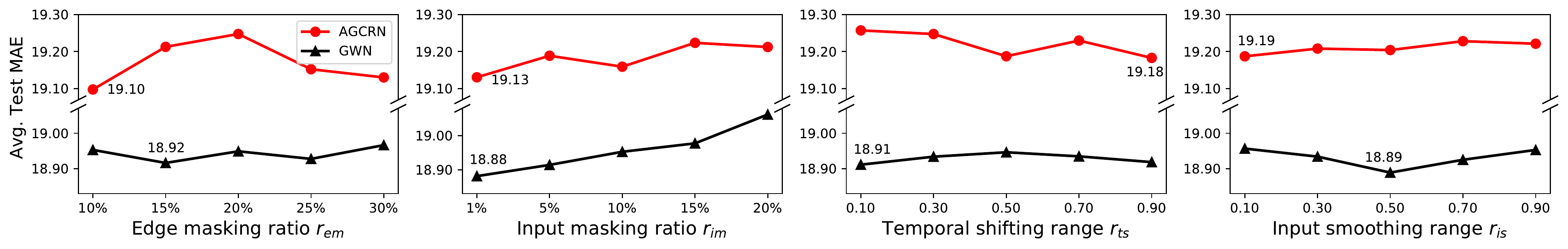}
         \vspace{-1.5em}
         \caption{PEMS-04}
         \label{fig:aug-04}
         \vspace{0.5em}
     \end{subfigure}
    %  \hfill
     \begin{subfigure}[b]{\textwidth}
         \centering
         \includegraphics[width=\textwidth]{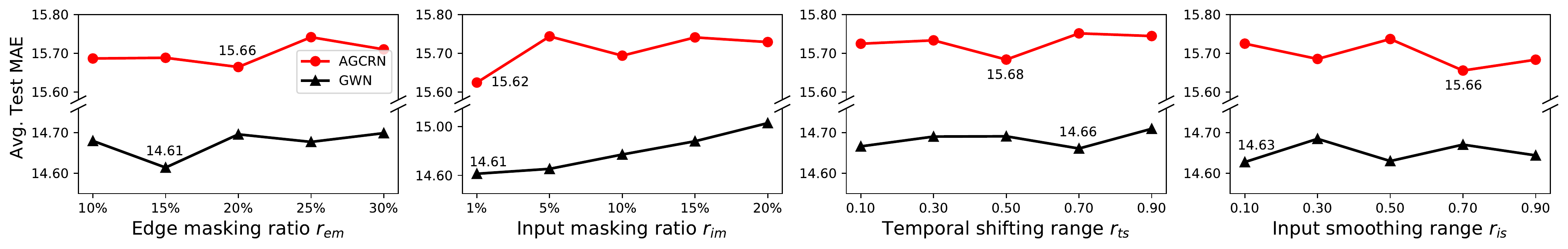}
         \vspace{-1.5em}
         \caption{PEMS-08}
         \label{fig:aug-08}
     \end{subfigure}
     \vspace{-1.5em}
    \caption{Effects of different augmentation methods on the PEMS-04 and PEMS-08 datasets.}
    \label{fig:aug}
    \vspace{-0.5em}
\end{figure*}

Moreover, we can see that in general, the graph-level contrast performs better than the node-level contrast. We speculate the reasons as follows. In the joint learning setting, nodes are guided to perform forecasting tasks, which results in a certain distribution in the latent space. Meanwhile, nodes are enforced to distinguish other nodes by the contrastive objective. In light of these factors, it is non-trivial to learn a powerful contrastive component, leading to minor improvements to the predictive performance. In contrast, it might be easier to distinguish patterns at the graph level. Also, since the number of neighbors to a certain node in these traffic datasets is usually small, it is more important and effective to utilize the graph-level contrast to provide global contexts to each node.

% different node have different behaviors
% node mix together, like have high degree, 
% calculate graph stat, in tradition graph, diameter is small, but in ours, diameter is large, nodes are far away from each other, gnn propagation only capture local information
% visualization the graph rep, find pattern, like 9 am on different date are closer, but in node rep, not such pattern
% compared to ssl on graph, they typically use node rep, but graph is better in our case, because the nature of graph is very different

Another important observation in Table \ref{tab:perf} is that contrastive methods achieve larger improvements for long-term predictions. To investigate it, we use GWN as the base model and randomly select a sensor (\#200) from PEMS-04 for a case study. Figure \ref{fig:case} visualizes the 60 minutes-ahead prediction results against the ground truth on a snapshot of the test data. It can be seen that GWN w/ JL-graph outperforms the counterpart, especially at the sudden change (see the red rectangle). A plausible explanation is that model has learned discriminative representations by receiving signals from the contrastive task during training, hence it can successfully distinguish some distinct patterns (like sudden changes).

% That is because STGCL can successfully distinguish the unique patterns (like sudden changes) by receiving supervision signals provided from the contrastive task.
% the curve of STGCL is more close to the ground truth data, which indicates the robustness that STGCL can offer; STGCL can predict the sudden change points more accurately than the base model.

\subsection{Ablation Studies (Q3 \& Q4)}
In this section, we set the joint learning scheme and the graph-level contrast as default due to their superior performance against their counterparts. Then we select one CNN-based model (GWN) and one RNN-based model (AGCRN) as representatives to study the effects of data augmentation and negative filtering.
\label{sec:4.4}

\subsubsection{Effects of data augmentation}
\label{sec:4.4.1}
We show the effects of different augmentation methods by tuning the hyper-parameter related to that method in Figure \ref{fig:aug}. We highlight the best performance achieved by each model on each augmentation method in the figure. For input smoothing, the fixed entries $E_{is}$ is set to 20 after tuning within the range of \{16, 18, 20, 22\}.

From the figures, we have the following observations: 1) all the proposed data augmentations yield gains over the base models. 2) The best performance achieved by each augmentation method has little difference, though the semantics of each method is different (e.g., temporal shifting and input smoothing are performed in the time and frequency domain, respectively). This indicates that our framework is not sensitive to the semantics of the proposed augmentations. This finding is also different from the situation in the vision domain, where data augmentation can notably vary the semantics and has a significant impact on the performance of the contrastive methods. 3) Input masking is significantly affected by the perturbation magnitude, while other augmentation methods are relatively stable and have no obvious trend. The reason could be that masking a certain fraction of entries to zero perturbs input a lot, but the perturbations injected by other augmentations are considered reasonable by the models. 4) Input masking with a 1\% ratio achieves most of the best performance across the applied models and datasets. We thereby suggest using this setting in other datasets for initiatives.
% Except for input smoothing, for the other augmentation methods, the same model shares the same trend across datasets.

\begin{figure}[!t]
  \centering
  \includegraphics[width=\linewidth]{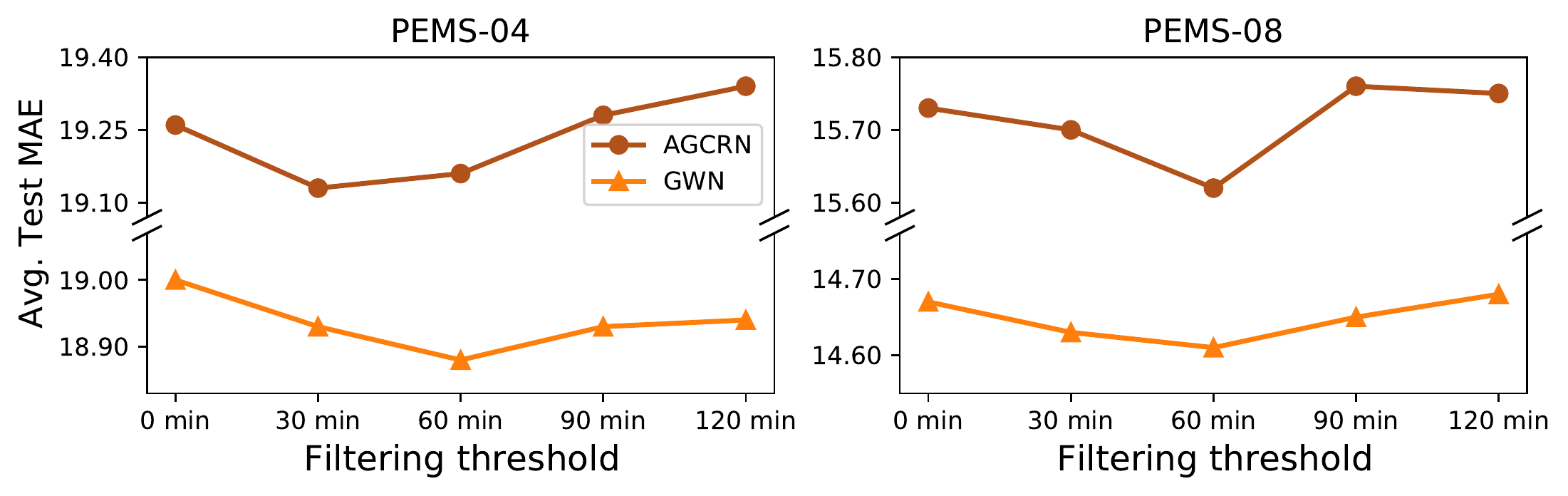}
%   \vspace{-1.5em}
  \caption{Effects of threshold $r_f$ on PEMS-04 and PEMS-08.}
  \label{fig:filter}
  \vspace{-0.5em}
\end{figure}

\subsubsection{Effects of negative filtering}
\label{sec:4.4.2}
We show the effects of the filtering threshold $r_f$ in Figure \ref{fig:filter}. Input masking with a 1\% ratio is used as the default augmentation setting. 
% loss term trade-off parameter $\lambda$ in Table \ref{tab:lambda}
% The results show that tuning $\lambda$ within \{0.01, 0.05, 0.1, 0.5, 1.0\} is sufficient to find the best value and the PEMS-04 dataset prefers a larger $\lambda$ than that on the PEMS-08 dataset.
According to the results, we can see that the filtering threshold $r_f$ works best when set to 30 or 60 minutes, which demonstrates the effectiveness of our simple solution. We also notice that when $r_f$ is set to 120 minutes, some results are worse than when $r_f$ is set to zero. The reason is that the threshold of 120 minutes filters out too many negatives, which eases the contrastive learning task. In this case, the model may not be able to learn useful discrimination knowledge that can benefit the forecasting task. To sum up, by filtering out the hardest negatives (i.e., the most semantically similar inputs), we enable the contrastive loss to focus on the \textit{true} negatives. On the other hand, filtering too many negatives makes the contrastive task meaningless and leads to performance degradation.

Our negative filtering operation is also more efficient in measuring the similarity of two temporal sequences than existing methods such as dynamic time warping \cite{berndt1994using} or Pearson correlation coefficient. The reason is that our method only needs to compare the starting time of each input and does not involve other computations.

% Furthermore, the negative filtering operation has the potential to be adapted in other time series-based contrastive methods [], as the time information is also available in their tasks.
% undesirable distinguish will accumulate

\subsection{Result Summaries \& Future Directions}
We summarize and discuss the experimental findings in this section, answering the four questions mentioned in Section \ref{sec:1}. Firstly, the joint learning scheme is able to consistently enhance base STG models, where the contrastive component works as a regularizer. Pretraining \& fine-tuning scheme optimizes different objective functions at the two stages, i.e., from self-supervision to forecasting loss, and we show that the knowledge derived from pretrained contrastive tasks does not improve forecasting performance. This finding is similar to the observation in the graph domain \cite{you2020does}.

Second, for the self-supervised tasks, we find that the node-level contrast outperforms the graph-level contrast in the pretraining scheme, but performs worse in the joint learning setting. This indicates that the auxiliary task cannot conflict with the main (forecasting) task in joint learning. In addition, since masked image modeling has surpassed the performance achieved by contrastive methods recently \cite{bao2022beit, he2022masked}, a future study may focus on acquiring additional training signals from masked modeling.

Third, we empirically show that the model is not sensitive to the semantics and the perturbation magnitude (only except for input masking) of the proposed augmentations. Motivated by the empirical results from SimCSE \cite{gao2021simcse}, a future study may conduct to apply a single dropout as the augmentation. Moreover, a possible direction for improving performance is to leverage adaptive data augmentation techniques \cite{zhu2021graph} to identify the importance of nodes/edges, thus treating them differently when generating augmented views.

Last, we should carefully select the negatives, i.e., the negatives should not be too similar/dissimilar to the anchor point. Designing a filtering method that dynamically assigns weights to hard negatives rather in a heuristic-based way could be one future direction.

\section{Related Work}
\subsection{Deep Learning for STG Forecasting}
STG forecasting is a typical problem in smart city efforts, facilitating a wide range of applications \cite{zheng2014urban}. In recent years, deep neural networks have become the dominant class for modeling STG \cite{han2021dynamic, xia20213dgcn, shi2020predicting, zhang2020semi, huang2019stgat}. Generally, they integrate graph convolutions with either CNNs or RNNs to capture the spatial and temporal dependencies in STG. As a pioneering work, DCRNN \cite{li2018diffusion} considers traffic flow as a diffusion process and proposes a novel diffusion convolution to capture spatial correlations. To improve training speed, GWN \cite{wu2019graph} adopts a complete convolutional structure, which combines graph convolution with dilated casual convolution operation. In addition, attention mechanisms \cite{liang2018geoman,guo2019attention} are proposed to capture dynamic inter- and inner-sensor correlations. The recent developments in this field show the following trends: making the adjacency matrix fully learnable \cite{bai2020adaptive}, and developing modules to jointly capture spatial and temporal dependencies \cite{song2020spatial}. However, these models are usually trained on datasets with limited instances, which may lead to overfitting problem and inferior generalization performance.
% Such forecasts have been widely applied in various applications, such as traffic flow/speed predictions and air quality predictions.

\subsection{Graph-based Contrastive Learning}
Recently, contrastive representation learning on graphs has attracted significant attention. According to a recent survey \cite{liu2021graph}, existing efforts can be categorized into cross-scale contrasting and same-scale contrasting.

Cross-scale contrasting refers to the scenario that the contrastive elements are in different scales. A representative work named DGI \cite{velickovic2019deep} contrasts between the patch and graph-level representations via mutual information (MI) maximization, thus gaining benefits from flowing global information to local representations. While MVGRL \cite{hassani2020contrastive} suggests a multi-view contrasting by using a diffused graph as a global view of the original graph, after which the MI is maximized in a cross-view and cross-scale manner.

For same-scale contrasting, the elements are in an equal scale, such as graph-graph contrasting. Regarding the definition of positive/negative pairs, the approaches can be further divided into context-based and augmentation-based. The context-based methods generally utilize random walks to obtain positives. Our work lies in the scope of augmentation-based methods, where the positives are generated by perturbations on nodes or edges. For example, GraphCL \cite{you2020graph} adopts four augmentations to form positive pairs and contrasts at graph level. GCA \cite{zhu2021graph} works at the node level and performs augmentations that are adaptive to the graph structure and attributes. IGSD \cite{zhang2020iterative} uses graph diffusion to generate augmented views and applies a teacher-student framework.

It can be seen that the current works mostly focus on developing contrastive methods for attributed graphs and pay less attention to more complex graphs, e.g., heterogeneous graphs and STG \cite{liu2021graph}. In this work, we present the first exploration on adapting contrastive learning to STG.

\section{Conclusion}
In this paper, we present the first systematic study on incorporating contrastive learning into STG forecasting. We first elaborate two potential schemes for integrating contrastive learning. Then we propose two feasible and efficient designs of contrastive tasks that are performed on the node or graph level. In addition, we introduce four types of STG-specific data augmentations to construct positive pairs and propose a rule-based strategy to alleviate an inherent shortcoming of the contrastive methods, i.e., ignoring input's semantic similarity. Our extensive evaluations on real-world STG benchmarks and different base models demonstrate that integrating graph-level contrast with the joint learning scheme achieves the best performance. We also provide rational explanations and insights for experimental results and discuss the future research directions.
% This work represents an initial attempt to exploit self-supervised learning for STG forecasting and opens up new research possibilities.

\section{Acknowledgements}
This research is supported by Singapore Ministry of Education Academic Research Fund Tier 2 under MOE's official grant number T2EP20221-0023. We thank all reviewers for their constructive suggestions to improve this paper. We also thank Hengchang Hu, Zhiyuan Liu, and Canwei Liu for their valuable feedback.

\bibliographystyle{ACM-Reference-Format}
\bibliography{sigspatial22}
\clearpage

\appendix

\section{Reproducibility}
Here, we detail the experimental settings to support reproducibility of the reported results in the paper. First, we present the dataset-related settings in Section \ref{sec:dataset-appendix}. Second, we provide the specific settings in Section \ref{sec:base-appendix} and \ref{sec:stgcl-appendix} that are used to obtain the results in Table \ref{tab:perf} and \ref{tab:perf-appendix}. Third, Section \ref{sec:twostage-appendix} shows the procedure of the pretraining \& fine-tuning experiments.

\subsection{Datasets}
\label{sec:dataset-appendix}
We conduct the experiments on the traffic datasets of PEMS-04\footnote{\label{note1}https://github.com/guoshnBJTU/ASTGCN-r-pytorch}, PEMS-08\footnotemark[3], and BAY\footnote{https://github.com/liyaguang/DCRNN}.
\begin{itemize}[leftmargin=*]
    \item PEMS-04 \cite{guo2019attention}: The dataset refers to the traffic flow data in the Bay Area. There are 307 sensors and the period of data ranges from Jan. 1 - Feb. 28, 2018.
    \item PEMS-08 \cite{guo2019attention}: The dataset contains traffic flow information collected from 170 sensors in the San Bernardino area from Jul. 1 - Aug. 31, 2016.
    \item BAY \cite{li2018diffusion}: This dataset contains traffic speed information from 325 sensors in the Bay Area. It has 6 months of data ranging from Jan. 1 - Jun. 30, 2017.
\end{itemize}

\begin{figure}[t]
  \centering
  \includegraphics[width=\linewidth]{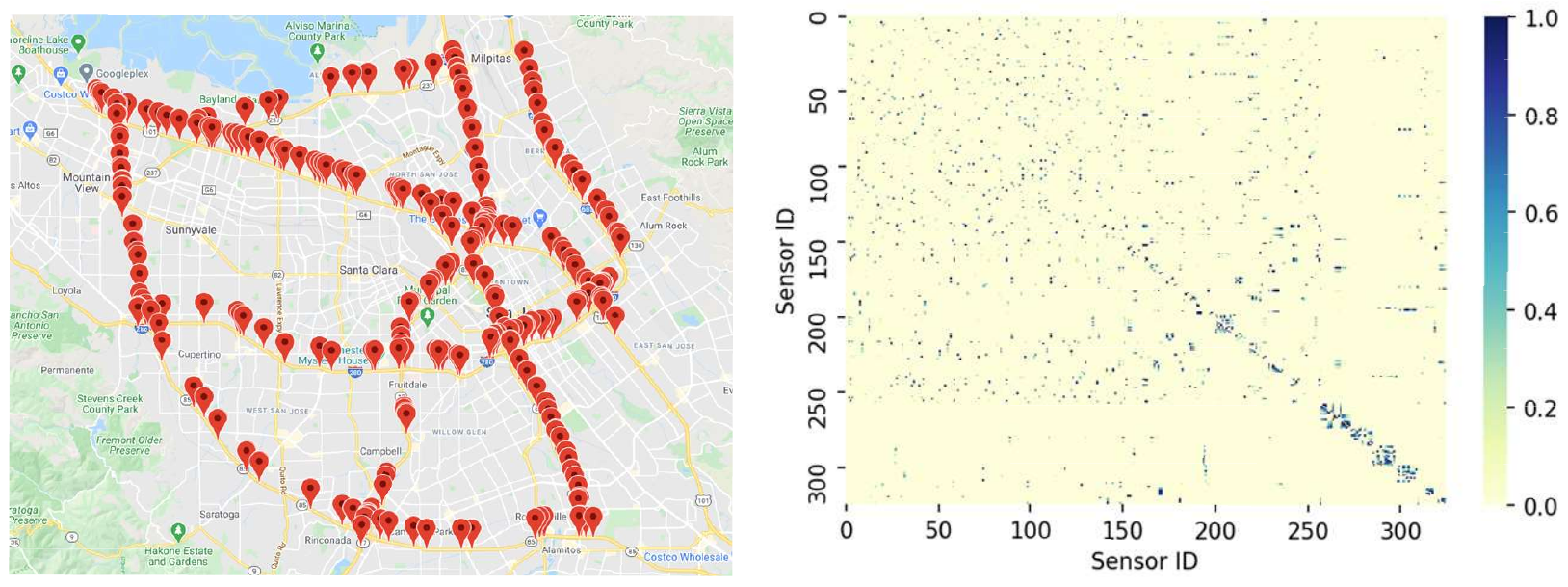}
  \caption{Visualizations of sensor distribution and the adjacency matrix on BAY.}
  \label{fig:sensor}
\end{figure}

We use one-hour (12-step) historical data to predict the future one-hour (12-step) values. The ``time of the day" is used as an auxiliary feature for the inputs. The datasets are split into three parts for training, validation, and testing with a ratio of 6:2:2 on PEMS-04 and PEMS-08, and 7:1:2 on BAY. The statistics of data partitions is in Table \ref{tab:partition}. Following \citet{li2018diffusion}, we build the adjacency matrix of the sensor graph in each dataset by using pairwise road network distances with a thresholded Gaussian kernel \cite{shuman2013emerging}. 
\begin{displaymath}
    W_{ij} = 
        \begin{cases}
            \exp(-\frac{{\rm dist}(v_i, v_j)^2}{\sigma^2}), & \text{if} \ {\rm dist}(v_i, v_j) \leqslant k \\
            0, & \text{otherwise}
        \end{cases}
\end{displaymath}
where $W_{ij}$ is the edge weight between sensor $v_i$ and sensor $v_j$, ${\rm dist}(v_i, v_j)$ denotes the road network distance between them, $\sigma$ is the standard deviation of distances, and $k$ is a threshold, which we set to 0.1. We provide a visualization of sensor distribution as well as the corresponding adjacency matrix on BAY in Figure \ref{fig:sensor}.

\begin{table}[h]
\centering
  \caption{Details of data partition.}
  \tabcolsep=3mm
  \label{tab:partition}
  \begin{tabular}{l|ccc}
    \shline
     Datasets & \#Training & \#Validation & \#Testing \\
    \hline
     PEMS-04 & 10,172 & 3,375 & 3,376 \\
     PEMS-08 & 10,690 & 3,548 & 3,549 \\
     BAY & 36,465 & 5,209 & 10,419  \\
    \shline
  \end{tabular}
\end{table}

% \paragraph{Evaluation Metrics}
% We adopt three kinds of metrics in regression tasks to evaluate the model performance, including mean absolute error (MAE), root mean squared error (RMSE), and mean absolute percentage error (MAPE). Formally, we have:

\begin{table*}[t]
\centering
\caption{Complete results of the four base models and base models w/ JL-graph on the test set of PEMS-04, PEMS-08, and BAY datasets. The first, second, and third rows of each setting indicate MAE, RMSE, and MAPE, respectively.}
\label{tab:perf-appendix}
\begin{tabular}{lccc|ccc|ccc}
\shline
\multirow{2}{*}{Methods} & \multicolumn{3}{c|}{PEMS-04} & \multicolumn{3}{c|}{PEMS-08} & \multicolumn{3}{c}{BAY} \\ \cline{2-10}
& 15 min & 30 min & 60 min & 15 min & 30 min & 60 min & 15 min & 30 min & 60 min \\ \hline

\multirow{3}{*}{GWN} & 18.20$\pm$.09 & 19.32$\pm$.13 & 21.10$\pm$.18 & 13.80$\pm$.05 & 14.75$\pm$.04 & 16.39$\pm$.09 & 1.31$\pm$.00 & 1.64$\pm$.01 & 1.97$\pm$.03 \\
& 29.15$\pm$.12 & 30.84$\pm$.15 & 33.38$\pm$.19 & 21.87$\pm$.04 & 23.71$\pm$.07 & 26.20$\pm$.11 & 2.75$\pm$.01 & 3.73$\pm$.04 & 4.55$\pm$.07 \\
& 12.67$\pm$.44 & 13.46$\pm$.38 & 15.05$\pm$.31 & 9.30$\pm$.42 & 9.88$\pm$.43 & 11.24$\pm$.51 & 2.71$\pm$.02 & 3.67$\pm$.03 & 4.62$\pm$.08 \\ \cline{2-10} 
 
\multirow{3}{*}{w/ JL-graph} & 17.93$\pm$.04 & 18.87$\pm$.04 & 20.40$\pm$.09 & 13.67$\pm$.04 & 14.61$\pm$.03 & 16.09$\pm$.05 & 1.29$\pm$.00 & 1.61$\pm$.00 & 1.88$\pm$.01 \\
& 28.86$\pm$.07 & 30.33$\pm$.07 & 32.47$\pm$.08 & 21.79$\pm$.04 & 23.66$\pm$.07 & 26.02$\pm$.12 & 2.72$\pm$.01 & 3.64$\pm$.01 & 4.34$\pm$.01 \\
& 12.36$\pm$.11 & 13.18$\pm$.25 & 14.61$\pm$.56 & 8.86$\pm$.16 & 9.74$\pm$.49 & 11.01$\pm$.58 & 2.70$\pm$.01 & 3.63$\pm$.02 & 4.46$\pm$.03 \\ \hline
 
\multirow{3}{*}{MTGNN} & 18.32$\pm$.05 & 19.10$\pm$.05 & 20.39$\pm$.09 & 14.36$\pm$.06 & 15.34$\pm$.10 & 16.91$\pm$.16 & 1.34$\pm$.02 & 1.66$\pm$.01 & 1.94$\pm$.02 \\
& 29.62$\pm$.13 & 31.15$\pm$.14 & 33.18$\pm$.16 & 22.47$\pm$.07 & 24.34$\pm$.08 & 26.75$\pm$.14 & 2.81$\pm$.01 & 3.76$\pm$.02 & 4.48$\pm$.03 \\
& 12.65$\pm$.20 & 13.14$\pm$.21 & 14.10$\pm$.21 & 9.38$\pm$.44 & 10.03$\pm$.25 & 11.50$\pm$.65 & 2.84$\pm$.09 & 3.74$\pm$.07 & 4.59$\pm$.08 \\ \cline{2-10}
 
\multirow{3}{*}{w/ JL-graph} & 17.99$\pm$.03 & 18.72$\pm$.05 & 19.88$\pm$.07 & 14.04$\pm$.05 & 14.90$\pm$.05 & 16.23$\pm$.08 & 1.32$\pm$.00 & 1.63$\pm$.01 & 1.89$\pm$.01 \\
& 29.03$\pm$.08 & 30.34$\pm$.13 & 32.08$\pm$.13 & 22.16$\pm$.08 & 23.99$\pm$.10 & 26.23$\pm$.16 & 2.83$\pm$.01 & 3.76$\pm$.02 & 4.38$\pm$.02 \\
& 12.35$\pm$.10 & 12.92$\pm$.25 & 13.93$\pm$.52 & 9.16$\pm$.18 & 9.70$\pm$.17 & 10.49$\pm$.09 & 2.79$\pm$.02 & 3.68$\pm$.04 & 4.42$\pm$.04 \\ \hline
 
\multirow{3}{*}{DCRNN} & 19.99$\pm$.11 & 22.40$\pm$.19 & 27.15$\pm$.35 & 15.23$\pm$.15 & 16.98$\pm$.25 & 20.27$\pm$.41 & 1.34$\pm$.03 & 1.71$\pm$.05 & 2.09$\pm$.08 \\
& 31.49$\pm$.16 & 34.97$\pm$.29 & 41.64$\pm$.49 & 23.74$\pm$.21 & 26.65$\pm$.34 & 31.45$\pm$.55 & 2.83$\pm$.07 & 3.89$\pm$.08 & 4.85$\pm$.16 \\
& 13.48$\pm$.05 & 15.17$\pm$.12 & 18.74$\pm$.21 & 9.63$\pm$.07 & 10.73$\pm$.10 & 12.90$\pm$.21 & 2.81$\pm$.06 & 3.90$\pm$.10 & 5.05$\pm$.23 \\ \cline{2-10} 
 
\multirow{3}{*}{w/ JL-graph} & 19.82$\pm$.08 & 22.07$\pm$.12 & 26.51$\pm$.21 & 15.19$\pm$.11 & 16.89$\pm$.19 & 20.09$\pm$.34 & 1.32$\pm$.00 & 1.68$\pm$.00 & 2.05$\pm$.00 \\
& 31.26$\pm$.10 & 34.53$\pm$.17 & 40.85$\pm$.30 & 23.71$\pm$.14 & 26.61$\pm$.24 & 31.30$\pm$.40 & 2.79$\pm$.00 & 3.83$\pm$.01 & 4.76$\pm$.01 \\
& 13.44$\pm$.04 & 15.05$\pm$.06 & 18.44$\pm$.13 & 9.59$\pm$.03 & 10.70$\pm$.08 & 12.82$\pm$.19 & 2.77$\pm$.01 & 3.82$\pm$.04 & 4.92$\pm$.03 \\ \hline
 
\multirow{3}{*}{AGCRN} & 18.53$\pm$.03 & 19.43$\pm$.06 & 20.72$\pm$.03 & 14.58$\pm$.07 & 15.71$\pm$.07 & 17.82$\pm$.11 & 1.37$\pm$.00 & 1.69$\pm$.01 & 1.99$\pm$.01 \\
& 29.70$\pm$.13 & 31.29$\pm$.16 & 33.26$\pm$.13 & 22.75$\pm$.09 & 24.70$\pm$.13 & 27.76$\pm$.21 & 2.88$\pm$.01 & 3.85$\pm$.01 & 4.59$\pm$.02 \\
& 12.87$\pm$.44 & 13.23$\pm$.29 & 13.97$\pm$.14 & 9.61$\pm$.17 & 10.64$\pm$.49 & 12.17$\pm$.55 & 2.94$\pm$.02 & 3.85$\pm$.02 & 4.67$\pm$.02 \\ \cline{2-10} 
 
\multirow{3}{*}{w/ JL-graph} & 18.31$\pm$.04 & 19.17$\pm$.06 & 20.39$\pm$.03 & 14.51$\pm$.05 & 15.56$\pm$.06 & 17.51$\pm$.10 & 1.35$\pm$.00 & 1.67$\pm$.01 & 1.96$\pm$.01 \\
& 29.69$\pm$.20 & 31.23$\pm$.23 & 33.05$\pm$.18 & 22.69$\pm$.07 & 24.57$\pm$.12 & 27.47$\pm$.13 & 2.85$\pm$.01 & 3.80$\pm$.02 & 4.56$\pm$.02 \\
& 12.63$\pm$.18 & 13.16$\pm$.25 & 13.96$\pm$.13 & 9.61$\pm$.29 & 10.24$\pm$.28 & 11.47$\pm$.17 & 2.90$\pm$.02 & 3.79$\pm$.04 & 4.62$\pm$.08 \\ 
\shline
\end{tabular}
\end{table*}

\subsection{Settings of the Base Models}
\label{sec:base-appendix}
We apply four base models that belong to the classes of CNN-based (GWN, MTGNN) and RNN-based methods (DCRNN, AGCRN). We basically employ the default settings of base models from their public source code. The common settings of the four models: Adam optimizer is applied, batch size is 64, and epochs are 100. The other settings of each model are listed as follows.

\begin{itemize}[leftmargin=*]
    \item GWN \cite{wu2019graph}: In this model, a spatio-temporal layer is constructed by a gated temporal convolution layer and a graph convolution layer. To cover the input sequence, eight spatio-temporal layers are applied and the dilation factors of each layer are 1, 2, 1, 2, 1, 2, 1, 2. The diffusion step and dropout rate in each graph convolution layer are set to 2 and 0.3. The hidden dimension size is 32. The learning rate, weight decay, and gradient clipping threshold are set to $1e^{-3}$, $1e^{-4}$, and 5.
    \item MTGNN \cite{wu2020connecting}: This model leverages three temporal convolution modules with the dilation factor 1, and three graph convolution modules to capture spatio-temporal correlations. The depth and the retain ratio of a mix-hop propagation layer (in graph convolution module) are set to 2 and 0.05. The saturation rate of the activation function from the graph learning layer is set to 3. The size of node embeddings is 40. Besides, dropout with 0.3 is applied after each temporal convolution layer. Layernorm is applied after each graph convolution layer. The learning rate, l2 penalty, and the threshold of gradient clipping are set to $1e^{-3}$, $1e^{-4}$, and 5. Curriculum learning strategy is turned off.
    \item DCRNN \cite{li2018diffusion}: This method applies an encoder-decoder architecture. Both encoder and decoder contain two recurrent layers and the number of hidden units is 64. The maximum steps of randoms walks is set to 2. The initial learning rate is $1e^{-2}$ and decreases to $\frac{1}{10}$ every 20 epochs starting from the 10th epochs. The threshold of gradient clipping is 5. For scheduled sampling, the probability decay is $\epsilon_i = \frac{\mu}{\mu + \text{exp}(i/\mu)}$, where $i$ is the number of training iterations and $\mu$ is the parameter to control the speed of convergence. $\mu$ is set to 2,000 on all datasets.
    \item AGCRN \cite{bai2020adaptive}: This approach uses a two-layer RNN to encoder the information and one linear layer to predict the future. The hidden units for all the AGCRN cells are set to 64. The embedding dimension is set as 10 for BAY and PEMS-04, and 2 for PEMS-08. The learning rate is $3e^{-3}$. Non-mentioned optimization tricks such as gradient clipping, are not applied.
\end{itemize}
Note that our reproduced results for the base models are close to, and some are even better than, the results reported in the original papers.

\subsection{Settings of STGCL}
\label{sec:stgcl-appendix}
For the settings described in Section \ref{sec:base-appendix}, STGCL uses the same, except that we don’t apply weight decay, as this will introduce another term in the loss function. 

For STGCL's specific settings, the temperature $\tau$ is set to 0.1 after tuning within the range of \{0.05, 0.1, 0.15, 0.2, 0.25\}. The augmentation method is fixed to input masking with a 1\% ratio, since we empirically find that it generally performs best as shown in Section \ref{sec:4.4.1}. The other two important hyper-parameters are $\lambda$ and $r_f$. The applied values of them are in Table \ref{tab:hyper}. We can observe that tuning $\lambda$ within \{0.01, 0.05, 0.1, 0.5, 1.0\} is sufficient to find the best value, and $r_f$ can set to 60 minutes as the default value.

\begin{table}[h]
\centering
\small
\tabcolsep=1.5mm
\caption{Values of the two important hyper-parameters (loss term trade-off parameter $\lambda$ and negative filtering threshold $r_f$) used in each STGCL's reported result in Table \ref{tab:perf-appendix}.}
\label{tab:hyper}
% \resizebox{\linewidth}{!}{
\begin{tabular}{l|cc|cc|cc}
\shline
\multirow{2}{*}{Methods} & \multicolumn{2}{c|}{PEMS-04} & \multicolumn{2}{c|}{PEMS-08} & \multicolumn{2}{c}{BAY} \\ \cline{2-7} 
 & $\lambda$ & $r_f$ & $\lambda$ & $r_f$ & $\lambda$ & $r_f$ \\ \hline
GWN w/ JL-node & 0.05 & 60 min & 0.05 & 60 min & 0.05 & 60 min \\
GWN w/ JL-graph & 0.5 & 60 min & 0.05 & 60 min & 1.0 & 60 min \\ \hline
MTGNN w/ JL-node & 0.01 & 60 min & 0.01 & 60 min & 0.5 & 60 min \\
MTGNN w/ JL-graph & 0.5 & 60 min & 1.0 & 60 min & 0.1 & 60 min \\ \hline
DCRNN w/ JL-node & 0.5 & 60 min & 0.1 & 60 min & 0.05 & 60 min \\
DCRNN w/ JL-graph & 0.1 & 60 min & 0.05 & 60 min & 0.01 & 60 min \\ \hline
AGCRN w/ JL-node & 0.1 & 60 min & 0.05 & 60 min & 0.1 & 60 min \\
AGCRN w/ JL-graph & 0.5 & 30 min & 0.1 & 60 min & 0.01 & 60 min \\ \shline
\end{tabular}
% }
\end{table}

\subsection{Settings of Two-stage Training}
\label{sec:twostage-appendix}
We use two models (GWN and AGCRN) to perform the pretraining \& fine-tuning experiments. The model-related configurations are used as the same in Section \ref{sec:base-appendix}.

According to the typical pretraining procedure \cite{you2020graph}, we first train the STG encoder of the applied model with a contrastive objective. Based on the experience in Section \ref{sec:stgcl-appendix}, we use input masking with a 1\% ratio as the augmentation method and set the temperature $\tau$ and filtering threshold $r_f$ to 0.1 and 60 minutes respectively. We optimize the encoder by Adam optimizer for a maximum of 100 epochs and use an early stop strategy with the patience of 10 by monitoring the training loss.

After training, we select the checkpoint with the lowest training loss to perform the downstream task. Specifically, we fine-tune the pretrained encoder with an untrained decoder to predict the future. We optimize the encoder and decoder by Adam optimizer, setting the learning rate to $1e^{-4}$ and $1e^{-3}$, respectively.

\end{document}